\documentclass[10pt,twocolumn,letterpaper]{article}

\usepackage{cvpr}              

\usepackage{xcolor}
\usepackage{amsmath}
\usepackage{amssymb}
\usepackage{array}
\usepackage{multirow}
\usepackage{color}
\usepackage{colortbl}
\usepackage{framed}
\usepackage{bm}
\usepackage{bbm}
\usepackage{xspace}
\usepackage{enumitem}
\usepackage{algorithm}
\usepackage{listings}
\usepackage{url}
\usepackage{booktabs}
\usepackage{pifont} 
\usepackage{subcaption} 
\usepackage{comment}

\DeclareMathOperator{\diag}{diag}

\usepackage{environ}         
\usepackage{etoolbox}        
\usepackage{graphicx}        

\usepackage[accsupp]{axessibility}  
\usepackage{mathtools}
\usepackage{amsmath,amsthm,amsfonts,bm}
\theoremstyle{definition}
\newtheorem{definition}{Definition}[section]

\newcommand{\mcl}[1]{\mathcal{#1}}
\newcommand{\mbb}[1]{\mathbb{#1}}
\newcommand{\mbf}[1]{\mathbf{#1}}
\newcommand{\Poincare}{Poincar\'e\xspace}
\newcommand{\norm}[1]{\left\lVert#1\right\rVert}
\newcommand{\innerprod}[1]{\langle#1\rangle}
\newcommand{\bTilde}[1]{\Tilde{\boldsymbol{#1}}}

\newcommand{\Secref}[1]{(Section~\ref{#1})}
\newcommand{\Secrefs}[2]{(Section~\ref{#1},~\ref{#2})}

\newlength{\myl}
\let\origequation=\equation
\let\origendequation=\endequation

\RenewEnviron{equation}{
  \settowidth{\myl}{$\BODY$}                       
  \origequation
  \ifdimcomp{\the\linewidth}{>}{\the\myl}
  {\ensuremath{\BODY}}                             
  {\resizebox{\linewidth}{!}{\ensuremath{\BODY}}}  
  \origendequation
}

\definecolor{Gray}{gray}{0.9}
\definecolor{LightCyan}{rgb}{0.88,0.95,1}
\definecolor{blond}{rgb}{0.98, 0.94, 0.75}
\definecolor{green}{rgb}{0.0, 0.62, 0.42}
\definecolor{lightgray}{rgb}{0.83, 0.83, 0.83}
\definecolor{purple}{HTML}{E3DAF0}

\newcommand{\ours}{HySAC\xspace}

\newcommand{\tit}[1]{\smallbreak\noindent\textbf{#1.}}
\newcommand{\tinytit}[1]{\noindent\textbf{#1.}}

\definecolor{cvprblue}{rgb}{0.21,0.49,0.74}
\usepackage[pagebackref,breaklinks,colorlinks,allcolors=cvprblue]{hyperref}

\def\paperID{16174} 
\def\confName{CVPR}
\def\confYear{2025}

\title{Hyperbolic Safety-Aware Vision-Language Models}

\author{Tobia Poppi$^{*\hspace{0.02cm}1,2}$\quad Tejaswi Kasarla$^{*\hspace{0.02cm}3}$ \quad Pascal Mettes$^3$ \quad Lorenzo Baraldi$^1$ \quad Rita Cucchiara$^{1,4}$\\
$^1$University of Modena and Reggio Emilia, Italy \quad $^2$ University of Pisa, Italy \\ $^3$ University of Amsterdam, Netherlands \quad $^4$IIT-CNR, Italy\\
{\tt\small $^1$\{name.surname\}@unimore.it} \quad \tt\small $^2$ \{name.surname\}@phd.unipi.it  \quad \tt\small $^3$ \{initial.surname\}@uva.nl }

\begin{document}
\maketitle
\def\thefootnote{*}\footnotetext{Equal contribution}\def\thefootnote{\arabic{footnote}}
\begin{abstract}
Addressing the retrieval of unsafe content from vision-language models such as CLIP is an important step towards real-world integration. Current efforts have relied on unlearning techniques that try to erase the model’s knowledge of unsafe concepts. While effective in reducing unwanted outputs, unlearning limits the model's capacity to discern between safe and unsafe content. In this work, we introduce a novel approach that shifts from unlearning to an awareness paradigm by leveraging the inherent hierarchical properties of the hyperbolic space.
We propose to encode safe and unsafe content as an entailment hierarchy, where both are placed in different regions of hyperbolic space.
Our \ours, Hyperbolic Safety-Aware CLIP, employs entailment loss functions to model the hierarchical and asymmetrical relations between safe and unsafe image-text pairs. This modelling -- ineffective in standard vision-language models due to their reliance on Euclidean embeddings -- endows the model with awareness of unsafe content, enabling it to serve as both a multimodal unsafe classifier and a flexible content retriever, with the option to dynamically redirect unsafe queries toward safer alternatives or retain the original output. Extensive experiments show that our approach not only enhances safety recognition but also establishes a more adaptable and interpretable framework for content moderation in vision-language models. Our source code is available at: {\small{\url{https://github.com/aimagelab/HySAC}}}
\medskip

{\color{BrickRed} \noindent\textit{\textbf{Warning:} This paper features explicit sexual content and other material that some readers may find disturbing, distressing, or offensive.}}

\end{abstract}    
\section{Introduction}
\label{sec:intro}

Large-scale vision-language models (VLMs) have achieved remarkable successes in various applications, including cross-model retrieval~\cite{radford2021learning}, text-to-image and image-to-text generation~\cite{liu2024visual,rombach2022high} and various downstream tasks~\cite{materzynska2022disentangling,shen2021much, wang2021actionclip}. Popular VLMs like CLIP~\cite{radford2021learning} and ALIGN~\cite{jia2021scaling} leverage vast amounts of web-scraped image-text data to learn rich multimodal representations by aligning visual and textual modalities. However, most large-scale datasets sourced from the web contain unsafe or inappropriate content, such as violence, nudity, or hate speech~\cite{birhane2021multimodal,birhane2021large}. The presence of such content not only raises ethical concerns but also introduces risks for real-world applications~\cite{wolfe2023contrastive,birhane2024into, hamidieh2024identifying}, where exposure to or misuse of this material can lead to legal and societal repercussions. Birhane~\etal~\cite{birhane2024into} also show that the increasing dataset scale can exacerbate hateful and unsafe content, as identified in the now removed LAION-5B~\cite{schuhmann2021laion,thiel2023identifying}. Addressing the issue of unsafe content in VLMs is therefore of utmost importance to ensure responsible AI practices.

Recent efforts to mitigate unsafe content in vision-language models have led to the development of methods specifically designed for NSFW (Not Safe For Work) content removal. Most of these works~\cite{gandikota2023erasing,poppi2024removing} have focused on unlearning (\ie, removing) the knowledge of unsafe content from the models. As a recent example, Poppi \etal~\cite{poppi2024removing} develop a fine-tuned version of CLIP which unlearns toxic concepts by redirecting their embeddings towards safe regions, so that retrieval always produces safe content even when the model is prompted with unsafe inputs.

In contrast, we propose an approach for managing unsafe content in VLMs: emphasizing \textit{awareness} over \emph{unlearning}. Rather than hiding the flaws of VLMs by ignoring NSFW content, we aim to equip the VLMs with the ability to distinguish between safe and unsafe content. This in turn helps users of the model to expose or redirect NSFW content when necessary, a crucial step toward improving user agency, understanding, and interpretability~\cite{ehsan2024seamful}.

Inspired by recent hyperbolic vision-language models~\cite{desai2023hyperbolic,pal2024compositional}, we introduce a hyperbolic framework that leverages the geometric properties of hyperbolic space to separate safe and unsafe content effectively. Using a paired dataset of safe and unsafe image-text inputs~\cite{poppi2024removing}, we adjust the embeddings to create an entailment-based~\cite{ganea2018hyperbolic} structure. In this setup, safe concepts are positioned closer to the origin of the hyperbolic space, while unsafe concepts are mapped further away.
Specifically, we introduce a \textit{hyperbolic safe-to-unsafe entailment} mechanism that ensures safe content encompasses unsafe representations within conical regions, defining clear safety boundaries and \textit{safety traversals} to dynamically adjust query embeddings along the hyperbolic space to promote safe retrievals or, alternatively, expose relevant unsafe content when necessary. This framework not only organizes data into safe and unsafe radius-based regions but also enables controlled movement within the space, allowing retrievals to favor safety as required. Experiments demonstrate that \ours achieves clear improvements in safety awareness, retrieval performance, and NSFW content handling across multiple datasets, with robustness in both safe content redirection and controlled unsafe content accessibility.
\section{Related Work}
\label{sec:related}

\tit{Unlearning in vision-language models}
Unlearning concepts and content has recently received a lot of attention, empowered by the success of vision-language models.
Various approaches have been explored, such as full model retraining, fine-tuning, machine unlearning~\cite{cao2015towards,ginart2019making,golatkar2020eternal,poppi2024multi,poppi2024unlearning}, and differential privacy~\cite{golatkar2022mixed}.
Some of these efforts have focused on text-to-image models, with the goal of removing specific styles, concepts, or objects~\cite{kumari2023ablating,zhang2023forget}. 

A particular emphasis has been placed on removing NSFW content, which encompasses inappropriate, unsafe, or illegal material. Schramowsky \etal~\cite{schramowski2023safe} steer the generation away from NSFW areas, defined by a fixed set of concepts. The embedding of NSFW concepts is applied as negative guidance during the text-conditioning phase, acting as safety guidance. 
Gandikota~\etal~\cite{gandikota2023erasing} erase visual concepts using only their name, with negative guidance serving as a teacher. 
Poppi \etal~\cite{poppi2024removing} focus on removing NSFW concepts from CLIP-like models by fine-tuning the entire model using ViSU, a multimodal dataset containing Safe/NSFW Image/Texts quadruplets. Unlike earlier approaches, which target specific components, their method fine-tunes the entire vision-and-language model, making content removal applicable to downstream tasks.

We introduce a method for handling NSFW concepts in CLIP-like models by fine-tuning them in hyperbolic space. We exploit the hierarchical properties of hyperbolic geometry, yielding a clear advantage: the model becomes explicitly aware of whether content is safe or NSFW, rather than merely removing knowledge of unsafe content. Similarly to our approach, safe generation works~\cite{schramowski2023safe, brack2023mitigating}, underscore that a deeper understanding of (un-)safety can improve the control of harmful content generation.

\tit{NSFW concept detection}
A related area of research is the automatic detection of NSFW content. Several methods have been proposed for detecting NSFW and toxic language~\cite{cauteruccio2022extraction,hidayatullah2019adult,markov2023holistic}, primarily in social media contexts. DistilBERT~\cite{sanh2019distilbert} has emerged as a promising model for this task, especially when fine-tuned for identifying adult content. Detecting inappropriate language presents a significant challenge, and this complexity extends to visual content, where various techniques have been developed to detect NSFW imagery~\cite{gandhi2020scalable,nichol2021glide,birhane2021large}. In this domain, models like NudeNet~\cite{bedapudi2019nudenet} specialize in detecting nudity, while Q16~\cite{schramowski2022can} serves as a broader classifier, capable of identifying a wider range of NSFW content. However, identifying inappropriate visual content remains a complex task, given the challenges posed by subtle visual cues, lack of contextual information, and limited data availability.

While these detection methods focus solely on identifying unsafe content, they do not address the problem of retrieving relevant, safe alternatives when an unsafe input is detected. Our method makes it possible to jointly detect NSFW content and provide a mechanism to shift NSFW queries towards safe but relevant alternatives.

\tit{Hyperbolic learning}
\label{subsec:related-hyperbolic}
A key advantage of hyperbolic space is its inherent ability to represent hierarchical or tree-like structures with minimal distortion~\cite{nickel2017poincare,chamberlain2017neural,sala2018representation,nickel2018learning}. 
A comprehensive list of recent advancements is documented in surveys by Mettes \etal~\cite{mettes2024hyperbolic} and Peng \etal~\cite{peng2021hyperbolic}. Foundational works for building neural networks in hyperbolic space~\cite{ganea2018hyperbolic, ganea2018hyperbolicnn, becigneul2018riemannian, khrulkov2020hyperbolic, shimizu2020hyperbolic} led to the use of hyperbolic models across multiple modalities such as images~\cite{liu2020hyperbolic,atigh2022hyperbolic,ermolov2022hyperbolic,francohyperbolic,ge2023hyperbolic,van2023poincare}, text~\cite{tifrea2018poincar,zhu2020hypertext,dhingra2018embedding,le2019inferring}, graphs~\cite{liu2019hyperbolic,chami2020low,choudhary2021self,choudhary2024hyperbolic} and recommender systems~\cite{mirvakhabova2020performance,wang2021fully}.

Recent work has shown the strong potential of hyperbolic learning for vision-language models~\cite{desai2023hyperbolic,ibrahimi2024intriguing, pal2024compositional} and demonstrated that training with a loss function enforcing entailment cones~\cite{ganea2018hyperbolic} leads to the emergence of hierarchical structures between the embeddings. While Desai \etal~\cite{desai2023hyperbolic} enforce entailment structure \textit{across} modalities, Pal \etal~\cite{pal2024compositional} also explicitly enforce structure \textit{within} modalities by leveraging object-level compositions and nouns from text.
We take inspiration from these approaches and enforce entailment relations across safe and unsafe embeddings, creating an interpretable space for traversing from unsafe regions to safe regions in CLIP models. 
\section{Preliminaries}
\label{sec:preliminaries}

Throughout this work, we operate in hyperbolic space, a Riemannian manifold with a constant negative curvature. Following Desai \etal~\cite{desai2023hyperbolic}, we use the Lorentz model, as it is better equipped to deal with numerical instabilities associated with the \Poincare distance metric~\cite{nickel2018learning,lin2023hyperbolic}.

\begin{figure*}[!t]
    \centering
    \includegraphics[width=0.99\textwidth]{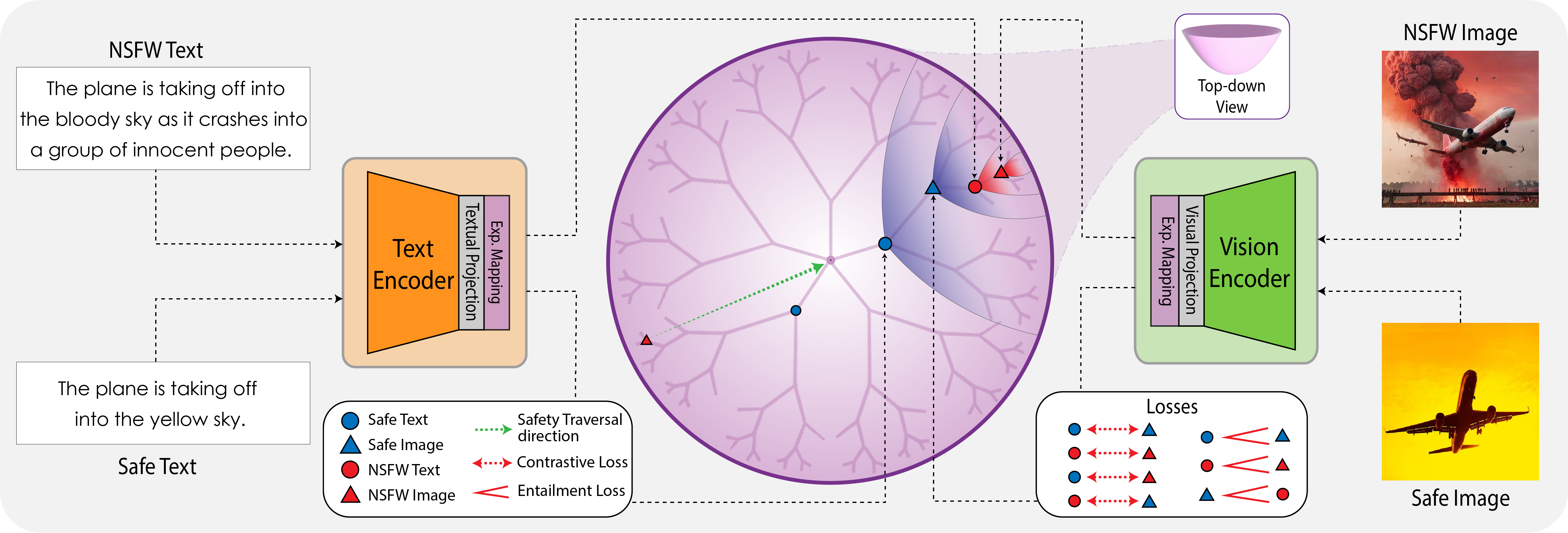}
    \caption{\textbf{Overview of our approach.} \ours builds a hyperbolic embedding that manages content safety through an entailment hierarchy. Unsafe text and images are projected to dedicated regions of hyperbolic space, allowing for safety-aware retrieval and classification.}
    \label{fig:model}
    \vspace{-.3cm}
\end{figure*}

The Lorentz model $\mbb{L}^n$ is an $n$-dimensional manifold in which points are represented on the upper sheet of a two-sheeted hyperboloid in $(n+1)$-dimensional Minkowski spacetime. Following terminology from general relativity, for each vector $\mbf{p} \in \mbb{R}^{n+1}$, we refer to the first dimension, the axis of symmetry, as the time-axis, denoted by $p_0$, and the remaining $n$-dimensions as the spatial components, denoted $\bTilde{p}$. The Lorentz model, $\mbb{L}^n = (\mcl{L}^n, \mathfrak{g}_{\boldsymbol{p}}^{\kappa})$ is given as
\vspace{-0.2cm}
\begin{equation}
    \mcl{L}^n \coloneq \left\{ \boldsymbol{p} \in \mbb{R}^{n+1} \colon \innerprod{\mbf{p}, \mbf{p}}_{\mcl{L}} = - \frac{1}{\kappa}, p_0 = \sqrt{1/\kappa + \norm{\bTilde{p}}^2}, \kappa > 0 \right\},
\end{equation}
where $-\kappa \in \mbb{R}$ denotes the curvature with the Riemannian metric $\mathfrak{g}_{\boldsymbol{p}}^{\kappa} = \diag(-1,1,\dots,1)$. The Lorentzian inner product $\langle .,. \rangle_{\mcl{L}}$ is induced by the metric tensor $\mathfrak{g}_{\boldsymbol{p}}^{\kappa}$
and is defined for $\mbf{p}, \mbf{q} \in \mbb{L}^n$ as 
\begin{equation}
    \innerprod{\mbf{p}, \mbf{q}}_{\mcl{L}} = -p_0 q_0 + \innerprod{\bTilde{p}, \bTilde{q}},
\end{equation}
where $\langle .,. \rangle$ is the Euclidean inner product. The Lorentzian inner product induces a norm on the Lorentzian space which can be written as $\norm{\mbf{p}}_{\mcl{L}} = \sqrt{\innerprod{\mbf{p}, \mbf{p}}_{\mcl{L}}}$. We now define the common hyperbolic operations in Lorentz space.

\begin{definition}[Lorentzian distance] The Lorentzian distance between two points in $\mbb{L}^n$ is the length of their shortest path (\textit{geodesic}) connecting them, computed as
\begin{equation}
d_{\mcl{L}}(\mbf{p}, \mbf{q}) = \sqrt{1/\kappa} \cdot \text{cosh}^{-1} ( - \kappa  \innerprod{\mbf{p}, \mbf{q}}_{\mcl{L}} ).
\label{eq:lorentzdist}
\end{equation}
\end{definition}
\vspace{-0.5cm}
In our work, we use the negative of Lorentzian distance to calculate similarities between multimodal inputs~\cite{desai2023hyperbolic, pal2024compositional}.

\begin{definition}[Exponential map] Since Lorentz space is a Riemannian manifold, it is locally Euclidean. This is best described through the tangent space $T_{\mbf{p}}\mbb{L}^n$, a first-order approximation of the Lorentzian manifold at a given point $\mbf{p} \in \mbb{L}^n$. The \textit{exponential map} then provides a means to project elements from the tangent space onto the hyperboloid. Given a point $\mbf{v} \in T_{\mbf{p}}\mbb{L}^n$, the exponential map is defined as $\text{exp}_{\mbf{p}}^{\kappa} \colon T_{\mbf{p}}\mbb{L}^n \rightarrow \mbb{L}^n$ with the expression
\begin{equation}
    \text{exp}_{\mbf{p}}^{\kappa} (\mbf{v}) = \text{cosh}(\sqrt{\kappa} \norm{\mbf{v}}_{\mbb{L}})\mbf{p} + \frac{\text{sinh}(\sqrt{\kappa} \norm{\mbf{v}}_{\mbb{L}})}{\sqrt{\kappa} \norm{\mbf{v}}_{\mbb{L}}} \mbf{v}.
    \label{eq:expmap}
\end{equation}
\end{definition}
\vspace{-0.4cm}
In practice, the reference point $\mbf{p}$ is set to the origin $\mbf{0}=(\sqrt{1/\kappa},0,...,0)^T$ on the hyperboloid, allowing  $\text{exp}_\mbf{0}^{\kappa}$ to project Euclidean vectors from the tangent space at $\mbf{0}$ directly onto the hyperboloid~\cite{khrulkov2020hyperbolic,desai2023hyperbolic}. In this work, the exponential map is used to project the outputs of the visual and textual encoders to a shared hyperbolic space.

\section{\ours: Hyperbolic Safety-Aware CLIP}
\label{sec:method}

\subsection{Problem formulation and objective}
\label{subsec:method-problem-formulation}

\tit{Problem setup}
Given a dataset $D=\{(I_i,T_i)\}_{i=1}^N$ of $N$ image-text pairs, vision-language models (\eg CLIP~\cite{radford2021learning}) align the visual and textual embeddings obtained from image and text encoders in a shared embedding space. Large-scale datasets employed for training such embedding spaces are often web-scraped and contain unsafe samples~\cite{birhane2021large}. For our problem setup, in order to differentiate between safe and unsafe content, we denote safe image-text pairs in $D$ as $(I_i,T_i)$ and unsafe image-text pairs as $(I_i^{\star}, T_i^{\star})$.

To make vision-language models aware of inappropriate contents, and enable them to avoid or redirect the representation of such content, we also require a dataset of quadruplets of safe and unsafe image-text pairs, denoted as $D^{\star} = \{(I_k,T_k,I_k^{\star}, T_k^{\star})\}_{k=1}^K$~\cite{poppi2024removing}.
This dataset is generated following the definition of NSFW content of Schramowski~\etal~\cite{schramowski2023safe} containing the following twenty categories: \textit{hate, harassment, violence, suffering, humiliation, harm, suicide, sexual, nudity, bodily fluids, blood, obscene gestures, illegal activity, drug use, theft, vandalism, weapons, abuse, brutality,} and \textit{cruelty}.
The dataset is constructed such that the unsafe image-text pairs $(I_k^{\star}, T_k^{\star})$ are specific cases or modified versions of the safe representations $(I_k, T_k)$.

\tit{Modelling relations in the hyperbolic space}
We first consider the relationship between text and image modalities in the embedding space. Like other hyperbolic vision-language models~\cite{desai2023hyperbolic, pal2024compositional}, we consider text as a general version of images to reflect the natural structure of partial order embeddings~\cite{vendrov2015order}. This is enforced in the embedding space by placing text embeddings closer to the origin and image embeddings farther away, defining a modality entailment relationship formally expressed as
\begin{equation}\label{eq:modality-inequality}
g_T(T_k) \ll g_I(I_k), \quad \text{and} \quad g_T(T_k^{\star}) \ll g_I(I_k^{\star}),
\end{equation}
where $g_I$ and $g_T$ denote the projections of images and text in the hyperbolic model and \( \ll \) indicates that one embedding is closer to the origin than another.
Differently from other models, we also impose a relationship between unsafe and safe pairs, by considering unsafe image-text pairs as specific cases of their safe counterparts within $D^{\star}$. Hence, we establish a safety entailment to segregate safe from unsafe content, as follows
\begin{equation}\label{eq:safety-inequality}
g_I(I_k) \ll g_T(T_k^{\star}).
\end{equation}
By satisfying these inequalities, \( g_I \) and \( g_T \) capture both the modal and safety hierarchies within the data, thereby endowing the embedding space with safety-aware properties.
To sum up, our objective is to model a hyperbolic vision-language model with the following inequality chain: 
\begin{equation}\label{eq:final-goal-safety}
           g_T(T_k) \ll g_I(I_k) \ll g_T(T_k^{\star}) \ll g_I(I_k^{\star}).\\              
\end{equation}
Below we outline our method to create safety-aware vision-language models in hyperbolic space.

\subsection{Hyperbolic safety learning}
To optimize Eq.~\ref{eq:final-goal-safety} for safety-aware vision-language models, we propose Hyperbolic Safety-Aware CLIP
to rearrange the embedding space to separate safe and unsafe regions. Our optimization consists of two components: (1) a hyperbolic safety contrastive component to align image-text pairs over a mini-batch and (2) a hyperbolic safety entailment to align safe and unsafe content. An overview is given in Figure~\ref{fig:model}.
\smallskip

\tit{Hyperbolic safety contrastive learning} CLIP~\cite{radford2021learning} and Safe-CLIP~\cite{poppi2024removing} rely on contrastive objectives to align and distribute the multimodal data. We utilize hyperbolic embeddings to align the visual and textual data. Specifically, we project the visual and textual embeddings from a pre-trained vision-language model onto a hyperboloid~\cite{desai2023hyperbolic} through an exponential map (Eq.~\ref{eq:expmap}). 
Let $f_I(.)$ and $f_T(.)$ represent any Euclidean encoders for image and text. Then, $g_I(I_k) = \text{exp}_{\mbf{0}}^{\kappa}(\alpha_{img} \cdot f_I(I_k))$ and $g_T(T_k) = \text{exp}_{\mbf{0}}^{\kappa}(\alpha_{txt} \cdot f_T(T_k))$ represent the hyperbolic representations of a safe image-text pair, $(I_k, T_k)$. Similarly,  $g_I(I_k^{\star}) = \text{exp}_{\mbf{0}}^{\kappa}(\alpha_{img} \cdot f_I(I_k^{\star}))$ and $g_T(T_k^{\star}) = \text{exp}_{\mbf{0}}^{\kappa}(\alpha_{txt} \cdot f_T(T_k^{\star}))$ represent the hyperbolic representations of an unsafe image-text pair $(I_k^{\star}, T_k^{\star})$. $\alpha_{img}$ and $\alpha_{txt}$ are learnable projection scalars.

To align representations in hyperbolic space, the similarity for the image-text and text-image contrastive loss is based on negative Lorentzian distance (Eq.~\ref{eq:lorentzdist}) between $g_I(.)$ and $g_T(.)$. We compute the hyperbolic safety contrastive loss over safe image-text pairs $(I_k,T_k)$ in a batch $B$ as
\begin{equation}
\label{eq:contr-safe-image-text-loss}
    L_{cont}^*(I,T) = - \sum_{i\in B} \log \frac{\text{exp}(d_\mcl{L}(g_I(I_i), g_T(T_i))/\tau)}{\sum_{k=1,k\neq i}^B \text{exp}(d_\mcl{L}(g_I(I_i), g_T(T_k))/\tau)},
\end{equation}
where $\tau$ denotes a temperature hyperparameter.  A similar contrastive loss is employed to preserve the multimodal structure between unsafe image-text pairs $(I_k^{\star}, T_k^{\star})$. Two additional contrastive losses between cross-safety modalities ensure the alignment of quadruplets in the embedding space. The final contrastive loss is formulated as
\begin{equation}\label{eq:contrastive-losses-all}
\begin{aligned}
    L_{\text{hSC}}(I,T,I^{\star},T^{\star}) = L_{cont}^{\star}(I,T) + L_{cont}^{\star}(I^{\star},T^{\star}) \\ 
    +L_{cont}^{\star}(I,T^{\star}) + L_{cont}^{\star}(I^{\star},T).
\end{aligned}  
\end{equation}

\tit{Hyperbolic safety entailment learning}
Hyperbolic entailment cones, introduced by Ganea \etal~\cite{ganea2018hyperbolic}, generalize partial ordered embeddings~\cite{vendrov2015order} to any Riemannian manifold. 
Entailment cones induce a partial order between concepts in a dataset $\mcl{X}$ such that for any pair $(\mbf{p}, \mbf{q}) \in \mcl{X}$, if $\mbf{p}$ is a subconcept of $\mbf{q}$, then $\mbf{q}$ entails $\mbf{p}$ within a conical region $\mathfrak{S_{\mbf{q}}}$ defined by $\mbf{q}$. For the Lorentz model, $\mbb{L}^n$, the half-aperture of each conical region $\mathfrak{S_{\mbf{q}}}$ is defined as~\cite{desai2023hyperbolic,le2019inferring}
\begin{equation}
    \omega(\mbf{q}) = \text{sin}^{-1}\left(\frac{2K}{\sqrt{\kappa}\norm{\bTilde{q}}} \right),
    \label{eq:half_aperture}
\end{equation}
where $-\kappa$ is the curvature of space, and the constant $K=0.1$ limits values near the origin~\cite{ganea2018hyperbolic}.
To preserve partial order between image-text relationships, we add entailment from safe text to safe image and from unsafe text to unsafe images, effectively implementing Eq.~\ref{eq:modality-inequality}. Specifically, a safe image $I_k$ must lie within the cone defined by its corresponding safe text, $T_k$, characterized by the half-aperture $\omega(T_k)$. Similarly, an unsafe image $I_k^{\star}$ must lie within the cone defined by its corresponding unsafe text $T_k^{\star}$. This is enforced through an entailment loss formulated for image-text representations by Le \etal~\cite{le2019inferring} and Desai~\etal~\cite{desai2023hyperbolic}, as
\begin{equation}
\label{eq:text_image_entail_loss}
\begin{aligned}
 L_{ent}^{\star}(I,T) = \text{max}(0,\phi(I_k,T_k)-\eta \omega(T_k)) \text{ and}\\
L_{ent}^{\star}(I^{\star},T^{\star}) = \text{max}(0,\phi(I_k^{\star},T_k^{\star})-\eta \omega(T_k^{\star})),
\end{aligned}
\end{equation}
where $\phi$ is the exterior angle (between lines $I_kT_k$ and $\mbf{0}T_k$ or between lines $I_k^{\star}T_k^{\star}$ and $\mbf{0}T_k^{\star}$) given by
\begin{equation}
\phi(I_k,T_k) = \text{cos}^{-1} \left( \frac{I_{k0}+T_{k0} \kappa \innerprod{I_k,T_k}_{\mcl{L}}}{\norm{\widetilde{T_k}}\sqrt{(\kappa \innerprod{I_k,T_k}_{\mcl{L}})^2-1}} \right).
\label{eq:ext_angle}
\end{equation}

Here, $\eta$ is a threshold for the half-aperture, $\omega(T_k)$~\cite{pal2024compositional}. Intuitively, the entailment loss $L_{ent}^{\star}$ penalizes images $I_k$ that lie outside the cone $\mathfrak{S}_{T_k}$ defined by their corresponding text caption $T_k$. 
Finally, to model the safety hierarchy, we enforce that safe concepts entail unsafe ones, reflecting that safe data are more general and unsafe data are more specific. Specifically, we enforce that a safe image $I_k$ entails an unsafe text $T_k^{\star}$, meaning that the unsafe text $T_k^{\star}$ must lie within the cone defined by the safe image $I_k$, characterized by the half-aperture $\omega(I_k)$. This gives us the safety-entailment defined in Eq.~\ref{eq:safety-inequality}. This is implemented as
\begin{equation}
L_{ent}^{\star}(T^{\star},I) = \text{max}(0,\phi(T^{\star}_k,I_k)-\eta \omega(I_k)).
\label{eq:entail_loss}
\end{equation}
The overall entailment loss to satisfy Eq.~\ref{eq:final-goal-safety} is defined as
\begin{equation}
    \label{eq:all-entailment-losses}
    L_{\text{hSE}}(I,T,I^{\star},T^{\star}) = L_{ent}^{\star}(I,T) + L_{ent}^{\star}(T^{\star},I) + L_{ent}^{\star}(I^{\star},T^{\star}).
\end{equation}

\tit{Combined loss function}
We integrate the contrastive with the entailment losses to obtain the total loss used to fine-tune the model on the dataset $D^{\star}$:
\begin{equation}
    \label{eq:final-loss}
    L (I,T,I^{\star}, T^{\star}) = L_{\text{hSC}}(I,T,I^{\star},T^{\star}) + L_{\text{hSE}}(I,T,I^{\star},T^{\star}).
\end{equation}
Our proposal allows the model to differentiate between safe and unsafe embeddings based on their distance from the origin. Safe content is closer to the center, while NSFW content is farther away. This geometric arrangement not only enables the model to detect unsafe content but also allows dynamic manipulation of embeddings. NSFW queries can be redirected toward the safe region, effectively retrieving outputs that prioritize safety and providing more precise control over content retrieval.

\subsection{Safety traversals and evaluation}
The hyperbolic safety-aware training results in a restructuring of the shared embedding space of the vision-language model. To obtain safe but relevant retrieval outputs from unsafe queries or vice-versa, we introduce a traversal mechanism to adjust query embeddings in hyperbolic space, enhancing their similarity with either safe or unsafe content, depending on the retrieval task. This traversal involves moving the query embeddings along the line connecting them to the origin of the hyperboloid, altering their hyperbolic distance from the root. By adjusting the embeddings' positions, we align them with regions in the embedding space that correspond to the desired content type.

\tit{Traversal Definition}
Given an embedding $\mbf{q}$, our method computes the distance from a predefined root feature $\mbf{r}$ in hyperbolic space using the Lorentzian distance function $d_{\mcl{L}}(\mbf{q}, \mbf{r})$.
For each type of content $X \in \{T, I, T^{\star}, I^{\star}\}$, we compute the mean distance $\mu_X$ from the root feature $\mbf{r}$ based on the distribution of each category. The boundary for each type is then defined as
\begin{equation} 
\tau_X = \mu_X + \tanh\left(\frac{\mu_X - \alpha}{\kappa}\right) + 1,
\end{equation}
where $\kappa$ is the negative curvature, and $\alpha$ is a constant set empirically to $0.8$. This shift accounts for the curvature of space, ensuring the boundaries are appropriately adjusted for effective traversal. Defining four bounds allows more nuanced control over traversal depending on the retrieval task. 
To retrieve a content type $X$, the query is moved along the Euclidean direction vector $\boldsymbol{v}_{\text{dir}} = \mbf{q} - \mbf{r}$ toward the root feature $\mbf{r}$ until it reaches the corresponding boundary $\tau_X$ (\eg $\tau_T$ for safe text). The target position $\mbf{q^*}$ is given as
\begin{equation}
\mbf{q^*} = \mbf{r} + \tau_X \cdot \frac{\boldsymbol{v}_{\text{dir}}}{\norm{\boldsymbol{v}_{\text{dir}}}}
\end{equation}
allowing the embeddings to be repositioned to match the target content type while maintaining semantic alignment.
\section{Experiments}
\label{sec:experiments}

\begin{table*}[t]
  \centering
  \setlength{\tabcolsep}{.45em}
  \resizebox{\linewidth}{!}{
  \begin{tabular}{lc ccc c ccc c ccc c ccc}
    \toprule
    & & \multicolumn{3}{c}{\textbf{Text-to-Image} ($T$-to-$I$)} & & \multicolumn{3}{c}{\textbf{Image-to-Text} ($I$-to-$T$)} & & \multicolumn{3}{c}{\textbf{Text-to-Image} ($T^{\star}$-to-$I\cup I^{\star}$)} & & \multicolumn{3}{c}{\textbf{Image-to-Text} ($I^{\star}$-to-$T\cup T^{\star}$)} \\
    \cmidrule{3-5} \cmidrule{7-9} \cmidrule{11-13} \cmidrule{15-17}
    \textbf{Model} & & R@1 & R@10 & R@20 & & R@1 & R@10 & R@20 & & R@1 & R@10 & R@20 & & R@1 & R@10 & R@20 \\
    \midrule
    CLIP~\cite{radford2021learning}& & 36.8 & 71.6 & 81.5 & & 39.8 & 74.2 & 83.5 & & 2.0 & 24.8 & 33.2 & & 4.6 & 32.9 & 40.6\\
    MERU~\cite{desai2023hyperbolic}& & 14.9 & 43.0 & 54.2 & & 14.7 & 42.3 & 53.8 & & 2.2 & 15.2 & 21.5 & & 4.4 & 22.6 & 29.4\\
    HyCoCLIP~\cite{pal2024compositional}& & 34.3 & 71.2 & 80.6 & & 34.4 & 71.3 & 82.2 & & 2.8 & 25.3 & 33.2 & & 8.2 & 37.8 & 45.7\\
    Safe-CLIP~\cite{poppi2024removing}& & 45.9 & 81.8 & 89.7 & & 45.3 & 82.3 & 89.8 & & 8.0 & 46.9 & 58.0 & & 19.1 & 62.9 & 71.1\\
    \midrule
    MERU$^\star$ & & 50.0 & 84.1 & 91.1 & & 51.2 & 85.3 & 92.3 & & 2.3 & 39.9 & 49.4 & & 5.7 & 47.9 & 54.7 \\
    HyCoCLIP$^\star$ & & 47.7 & 81.9 & 89.1 & & 46.7 & 82.7 & 90.4 & & 1.5 & 32.7 & 42.3 & & 6.9 & 45.2 & 53.6 \\

    \midrule
    \rowcolor{purple}
    \textbf{\ours} & & \textbf{49.8} & \textbf{84.1} & \textbf{90.7} & & \textbf{48.2} & \textbf{84.2} & \textbf{91.2} & & \textbf{30.5} & \textbf{62.8} & \textbf{71.8} & & \textbf{42.1} & \textbf{73.3} & \textbf{79.8}\\
    \bottomrule
  \end{tabular}
}
\vspace{-0.2cm}
  \caption{\textbf{Safe content retrieval performance on ViSU test set.} Across all tasks and recall rates, \ours improves over existing safety unlearning CLIP and hyperbolic CLIP models, highlighting that our approach is able to navigate unsafe image or text inputs towards relevant but safe retrieval outputs. $^\star$ CLIP fine-tuned in hyperbolic space on ViSU training set with MERU/HyCoCLIP losses.}
  \label{tab:retrieval}
  \vspace{-0.1cm}
\end{table*}

\begin{table*}[t]
  \centering
  \setlength{\tabcolsep}{.45em}
  \resizebox{\linewidth}{!}{
  \begin{tabular}{lc ccc c ccc c ccc c ccc}
    \toprule
    & & \multicolumn{3}{c}{\textbf{Text-to-Image} ($T^{\star}$-to-$I^{\star}$)} & & \multicolumn{3}{c}{\textbf{Image-to-Text} ($I^{\star}$-to-$T^{\star}$)} & & \multicolumn{3}{c}{\textbf{Text-to-Image} ($T^{\star}$-to-$I^{\star}\cup I$)} & & \multicolumn{3}{c}{\textbf{Image-to-Text} ($I^{\star}$-to-$T^{\star}\cup T$)} \\
    \cmidrule{3-5} \cmidrule{7-9} \cmidrule{11-13} \cmidrule{15-17}
    \textbf{Model} & & R@1 & R@10 & R@20 & & R@1 & R@10 & R@20 & & R@1 & R@10 & R@20 & & R@1 & R@10 & R@20 \\
    \midrule
    CLIP~\cite{radford2021learning} & & 73.1 & 94.9 & 97.6 & & 72.8 & 95.2 & 97.7 & & 68.4 & 92.3 & 95.9 & & 67.1 & 93.3 & 96.7\\
    MERU~\cite{desai2023hyperbolic} & & 29.4 & 62.4 & 72.2 & & 25.8 & 57.7 & 67.8 & & 23.5 & 54.0 & 64.3 & & 19.5 & 51.1 & 61.2\\
    HyCoCLIP~\cite{pal2024compositional} & & 69.5 & 93.1 & 95.8 & & 65.0 & 91.1 & 95.0 & & 63.7 & 89.7 & 93.7 & & 55.2 & 88.0 & 92.7\\
    Safe-CLIP~\cite{poppi2024removing} & & 58.0 & 86.2 & 91.4 & & 56.0 & 85.1 & 91.0 & & 47.7 & 80.0 & 85.8 & & 32.1 & 77.1 & 84.6\\
    \midrule
    \rowcolor{purple}
    \textbf{\ours} & & \textbf{81.4} & \textbf{98.4} & \textbf{99.4} & & \textbf{82.2} & \textbf{97.8} & \textbf{99.2} & & \textbf{81.1} & \textbf{98.4} & \textbf{99.4} & & \textbf{80.5} & \textbf{97.2} & \textbf{98.9}\\
    \bottomrule
  \end{tabular}
}
\vspace{-0.2cm}
  \caption{\textbf{Unsafe content retrieval performance on ViSU test set.} Akin to safe content retrieval, our approach performs best. This is a result of our objective, as we assign different content to different regions, enabling us to maintain valuable safety information.}
  \label{tab:retrieval-unsafe}
  \vspace{-0.3cm}
\end{table*}

\subsection{Training Details}
\label{subsec:implementation-details}
\tinytit{Datasets} Our experiments are mainly conducted on the ViSU dataset~\cite{poppi2024removing}, containing $165$k quadruplets of safe and unsafe image-text pairs.
We also evaluate our model on three real-world NSFW image datasets: NudeNet~\cite{bedapudi2019nudenet}, NSFW data source URLs\footnote{\href{https://github.com/EBazarov/nsfw_data_source_urls}{https://github.com/EBazarov/nsfw\_data\_source\_urls}} and SMID~\cite{crone2018socio}.

\tinytit{Baselines} Our safety comparisons include the original CLIP~\cite{radford2021learning} and the state-of-the-art Safe-CLIP~\cite{poppi2024removing}. CLIP was trained using a private dataset of $400$M image-text pairs~\cite{schuhmann2021laion} which has unsafe data~\cite{birhane2021large}. Safe-CLIP is finetuned on the ViSU dataset~\cite{poppi2024removing} to redirect unsafe content to safe correspondent one via contrastive losses and cosine similarities, aiming to unlearn NSFW concepts.

\tinytit{Models} Our visual and textual encoders are the same as CLIP~\cite{radford2021learning}, with VIT-L/14 as visual encoder, to maintain fair comparison to Safe-CLIP~\cite{poppi2024removing}. During training, both the visual and textual encoder are fine-tuned using low-rank decomposition~\cite{hu2021lora} with low-rank factor $r=16$.

\tinytit{Optimization} We use AdamW~\cite{loshchilov2018decoupled} with weight decay $0.2$ and $(\beta_1, \beta_2) = (0.9, 0.98)$. We disable weight decay for all gains, biases, and learnable scalars. The model is finetuned for $20$ epochs with batch size $256$. The maximum learning rate is $8\times10^{-4}$. We use mixed precision~\cite{micikevicius2018mixed} to accelerate training, except computing exponential map and losses for \ours in FP32 precision for numerical stability.

\tinytit{Initialization} We initialize image and text encoders akin to CLIP, along with pre-trained weights.
We initialize the softmax temperature as $\tau = 0.07$ and clamp it to a minimum value of $0.01$. For \ours, we initialize the learnable projection scalars $\alpha_{img} = \alpha_{txt} = 1/\sqrt{512}$, the curvature parameter $c = 1.0$ and clamp it in $[0.1, 10.0]$ to prevent training instability. All scalars are learned in logarithmic space as $\log(1/\tau)$, $\log(c)$, $\log(\alpha_{img})$ and $\log(\alpha_{txt})$.

Further details on the training setup are provided in the supplementary~\ref{sec:training-details-addl}.

\subsection{Experimental Results}
\label{subsec:results}
To assess the performance of our proposed model, \ours, we measure its safety awareness and its ability to handle unsafe content effectively, while retaining both safe and unsafe knowledge. In the supplementary~\ref{sec:additional-ablations}, we report the zero-shot generalization of our method.

\subsubsection{Safety retrieval comparison}
We evaluate our model on the capability of retrieving safe and unsafe items, in comparison to CLIP~\cite{radford2021learning} and Safe-CLIP~\cite{poppi2024removing}. We also provide a comparison to recent hyperbolic VLMs, namely MERU~\cite{desai2023hyperbolic} and HyCoCLIP~\cite{pal2024compositional}. Safe-CLIP and \ours fine-tuned from CLIP on ViSU~\cite{poppi2024removing}, MERU trained on RedCaps~\cite{desai2021redcaps}, and HyCoCLIP on GRIT~\cite{peng2023kosmos}. For a fair comparison, we additionally fine-tuned the CLIP model on the ViSU dataset in hyperbolic space, using MERU and HyCoCLIP\footnote{The box data needed for HyCoCLIP was extracted using Kosmos-2.} losses. All the models are evaluated on the ViSU test set.

Retrieval tasks are defined as text-to-image and image-to-text, where the goal is to find the most relevant counterpart for a given query.
Recall@K measures the fraction of queries where the correct item appears in the top-K retrieved results.
To assess the safe retrieval performance of \ours, we measure recall exclusively on safe content in both visual and textual elements ($T\text{-to-}I$ and $I\text{-to-}T$). This step is crucial to verify that the original CLIP model's retrieval capabilities are retained after finetuning in hyperbolic space with our training method. 
Then, to evaluate the safety-awareness capabilities of \ours, we introduce a distinct setup in which NSFW elements are used as queries, while the retrievable items include both safe and unsafe elements ($T^*\text{-to-}I \cup I^*$ and $I^*\text{-to-}T \cup T^*$). During these experiments, a retrieval is deemed correct only if the query retrieves its safe counterpart, thereby validating the model's ability to redirect unsafe queries towards safe items. 

When retrieving with \ours, the threshold $\tau$ for moving query embeddings is computed using the mean distance of safe embeddings from the origin, adjusting the query embedding towards the safe region. Results are reported in Table~\ref{tab:retrieval}, where we observe that \ours consistently improves over both unlearning and existing hyperbolic models and features the highest recalls across all settings and rates.
CLIP hyperbolic models finetuned on ViSU data (MERU$^\star$ and HyCoCLIP$^\star$) perform well on safe-only retrieval, while our method achieves high performance on both safe-only and unsafe-safe retrieval, due to our safety-aware design.

In Table~\ref{tab:retrieval-unsafe}, we instead perform analyses for unsafe content retrieval. First, this involves the text-to-image and image-to-text retrieval on only unsafe elements ($T^*\text{-to-}I^*$ and $I^*\text{-to-}T^*$). Second, instead, we perform retrieval by using unsafe elements as queries and both safe and unsafe items as retrievable items ($T^*\text{-to-}I^* \cup I$ and $I^*\text{-to-}T^* \cup T$), and deem the retrieval correct only if the query retrieves its corresponding unsafe one. This setup tests the model's ability to function as a content moderator and also showcases its capacity to provide user autonomy in content retrieval decisions. 
In these tests, the traversal mechanism in \ours uses adjusted parameters to move in the unsafe direction, targeting the retrieval of NSFW content. Here too, \ours achieves the best recall across all settings, demonstrating that \ours not only prioritizes safety by navigating away from NSFW content when required but also ensures that users can access NSFW content under controlled conditions, better than existing competitors.

\begin{figure}[tb]
  \centering
    \includegraphics[width=\linewidth]{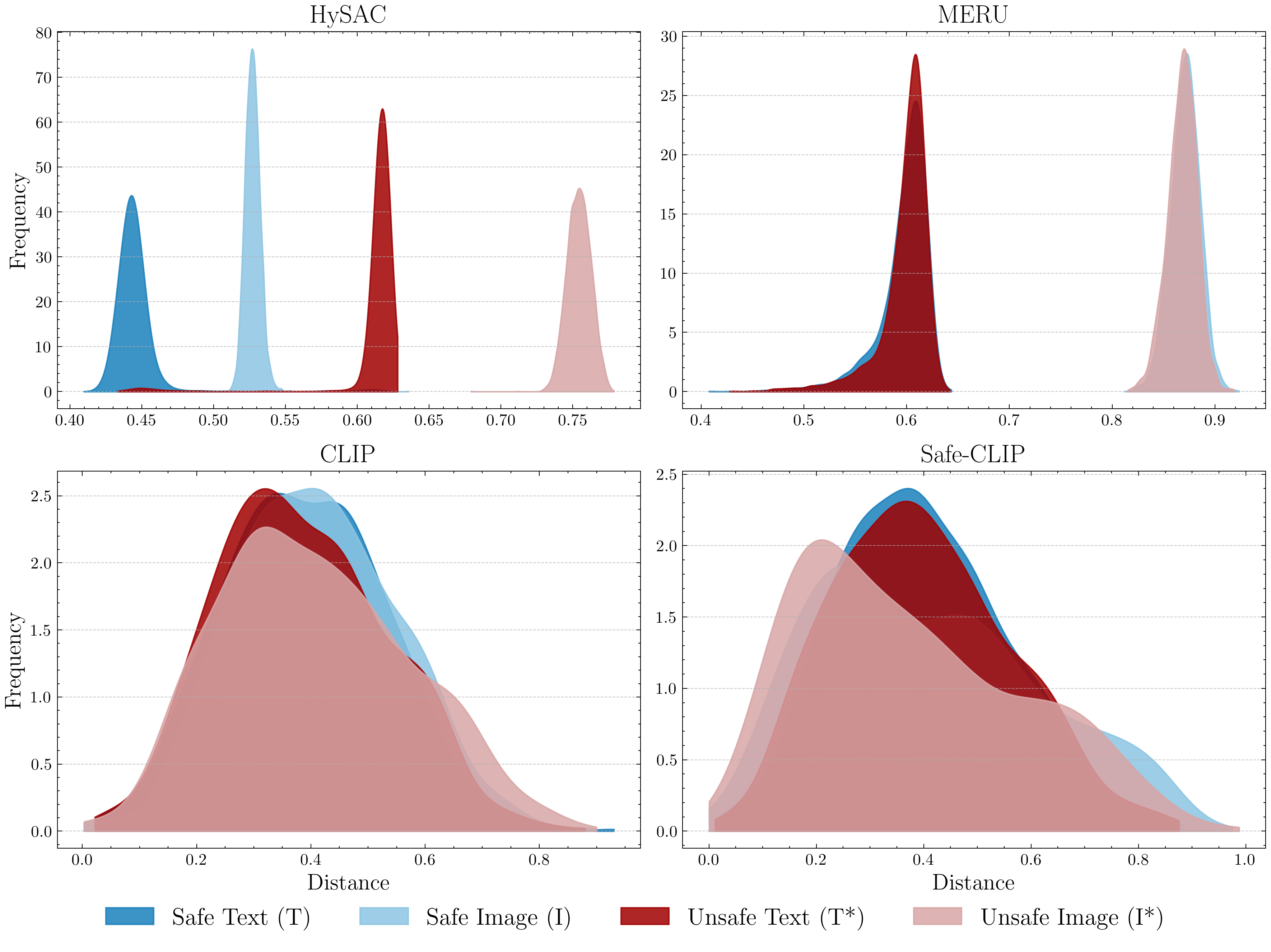}
    
  \vspace{-.2cm}
  \caption{\textbf{Distributions of embedding distances from the root}. We embed all ViSU training samples and visualize their distance distribution from the root. While CLIP and Safe-CLIP do not separate between texts and images, MERU does. \ours, instead, also differentiates between safe and unsafe content.}
  \label{fig:embedding_histograms}
  \vspace{-.2cm}
\end{figure}

\subsubsection{Analysis of \ours}
\begin{table}[t]
  \centering
  \setlength{\tabcolsep}{.22em}
  \resizebox{\linewidth}{!}{
  \begin{tabular}{lc cc c cc c cc c cc}
    \toprule
    & & \multicolumn{2}{c}{($T$-to-$I$)} & & \multicolumn{2}{c}{($I$-to-$T$)} & & \multicolumn{2}{c}{($T^{\star}$-to-$I\cup I^{\star}$)} & & \multicolumn{2}{c}{($I^{\star}$-to-$T\cup T^{\star}$)} \\
    \cmidrule{3-4} \cmidrule{6-7} \cmidrule{9-10} \cmidrule{12-13}
    \textbf{Model} & & R@1 & R@10 & & R@1 & R@10 & & R@1 & R@10 & & R@1 & R@10 \\
    \hline
    w/o Ent & & 52.3 & 84.9 & & 50.8 & 84.7 & & 4.1 & 49.0 & & 5.5 & 64.5 \\
    w/o S-Ent & & 51.0 & 84.2 & & 49.8 & 84.3 & & 1.4 & 39.1 & & 7.4 & 63.7 \\
    \rowcolor{purple}
    \textbf{\ours} & & 49.8 & 84.1 & & 48.2 & 84.2 & & \textbf{30.5} & \textbf{62.8} & & \textbf{42.1} & \textbf{73.3}\\
    \bottomrule
  \end{tabular}
}
\vspace{-0.2cm}
  \caption{\textbf{Ablation study on loss components}. We evaluate \ours against two ablations that remove loss components. Results are in the same setting of Table~\ref{tab:retrieval}.}
  \label{tab:ablation}
  \vspace{-0.5cm}
\end{table}

\tinytit{Assessing \ours embedding space}
Further, we validate the organization of the embedding space as outlined in Equation~\ref{eq:final-goal-safety}.
Our goal is to confirm that the embeddings for safe content are positioned closer to the origin of hyperbolic space, while those for unsafe content are further away, following the proposed hierarchy. Specifically, for each $X \in \{T, I, T^{\star}, I^{\star}\}$ of the training-set of $D^{\star}$, we compute the distances $d_{\mcl{L}}({X}, \mbf{r})$ from the root feature $\mbf{r}$.

A visualization is reported in Figure~\ref{fig:embedding_histograms}, where we show the distribution of embeddings in terms of their distance to the root feature. The comparative analysis is done across four different models: \ours, CLIP, Safe-CLIP, and MERU. For both the hyperbolic models the root feature is the origin of the hyperboloid. For the Euclidean models, since the origin does not lie on the hypersphere, the root is empirically estimated as the embedding that has the least distance from all embeddings of the training set, \ie~the $\ell_2$ normalization of the average of all embeddings. As it can be seen, the distribution clearly shows four peak distributions for \ours, each one representing one of the $X$ content types, elucidating the efficacy of our approach in maintaining a clear separation between safe, unsafe, textual, and visual content within the embedding space.

\tinytit{Ablation Study}
In Table~\ref{tab:ablation}, we validate the effectiveness of the key components in \ours, by comparing its full configuration with variants where specific losses are disabled. In particular, we employ one variant which only keeps contrastive losses (denoted as ``w/o Ent'') and one that omits the safety-entailment loss (``w/o S-Ent''). Results show that while removing these components slightly improves performance in scenarios involving only safe content, likely due to reduced spatial constraints, their absence significantly undermines the model’s effectiveness in dealing with unsafe content, especially in unsafe-to-safe retrieval. These results underscore the essential roles that both the modality- and safety-entailment losses play in enhancing the safety awareness of the proposed model.

\tinytit{Retrieval on real NSFW datasets} To further analyze the safety of \ours, we conduct retrieval tests using NudeNet~\cite{bedapudi2019nudenet}, NSFW data source URLs and SMID~\cite{crone2018socio}. The first two datasets primarily contain nudity and pornographic content, whereas SMID includes a broader range of inappropriate content, such as \textit{violence}, \textit{harm}, and \textit{discrimination}. We randomly select 1000 images from each dataset to serve as visual elements and use 5k NSFW captions from the ViSU test set. Both the image-to-text and text-to-image retrieval tasks also incorporate 10,000 randomly chosen retrievable safe items from LAION-400M~\cite{schuhmann2021laion}.

Results, displayed in Table~\ref{tab:t2i_nsfw}, contrast the performance of \ours with that of CLIP~\cite{radford2021learning} and Safe-CLIP~\cite{poppi2024removing}. We report the proportion of safe retrieval outputs when NSFW queries are used, highlighting the capability of \ours to enhance safety in retrieval results.
Traversing the embedding space towards safety prioritizes safer alternatives, which may adjust relevance in some cases. Due to the absence of datasets with real unsafe data and correlated safe alternatives, this aspect could not be evaluated.
Notably, \ours demonstrates improved performance in securing safer content compared to Safe-CLIP across most datasets for both text-to-image and image-to-text scenarios.

\begin{table}[t]
  \centering
  \setlength{\tabcolsep}{.22em}
  \resizebox{\linewidth}{!}{
  \begin{tabular}{lc ccc c ccc}
    \toprule
    & & \multicolumn{3}{c}{\textbf{\% Safe (Text-to-Image)}} & & \multicolumn{3}{c}{\textbf{\% Safe (Image-to-Text)}}\\
    \cmidrule{3-5} \cmidrule{7-9}
    \textbf{Model} & & NudeNet & NSFW URLs & SMID & & NudeNet & NSFW URLs & SMID \\
    \midrule
    CLIP & & 78.2 & 79.7 & 55.2 & & 33.3 & 44.0 & 59.1 \\
    Safe-CLIP & & 92.6 & 92.6 & \textbf{83.4} & & 75.2 & 76.4 & 65.6 \\
    \rowcolor{purple}
    \textbf{\ours} & & \textbf{96.2} & \textbf{93.9} & 80.1 & & \textbf{84.4} & \textbf{95.1} & \textbf{97.9} \\
    \bottomrule
  \end{tabular}
}
    \vspace{-.2cm}
  \caption{\textbf{Retrieval performance on real NSFW images.} Rate of safe images retrieved using unsafe prompts from the ViSU test set. The retrievable set includes safe and unsafe real images, with the latter from LAION-400M and the former from NSFW sources.}
  \label{tab:t2i_nsfw}
  \vspace{-.5cm}

\end{table}

\begin{figure*}[!t]
    \centering
    \includegraphics[width=\textwidth]{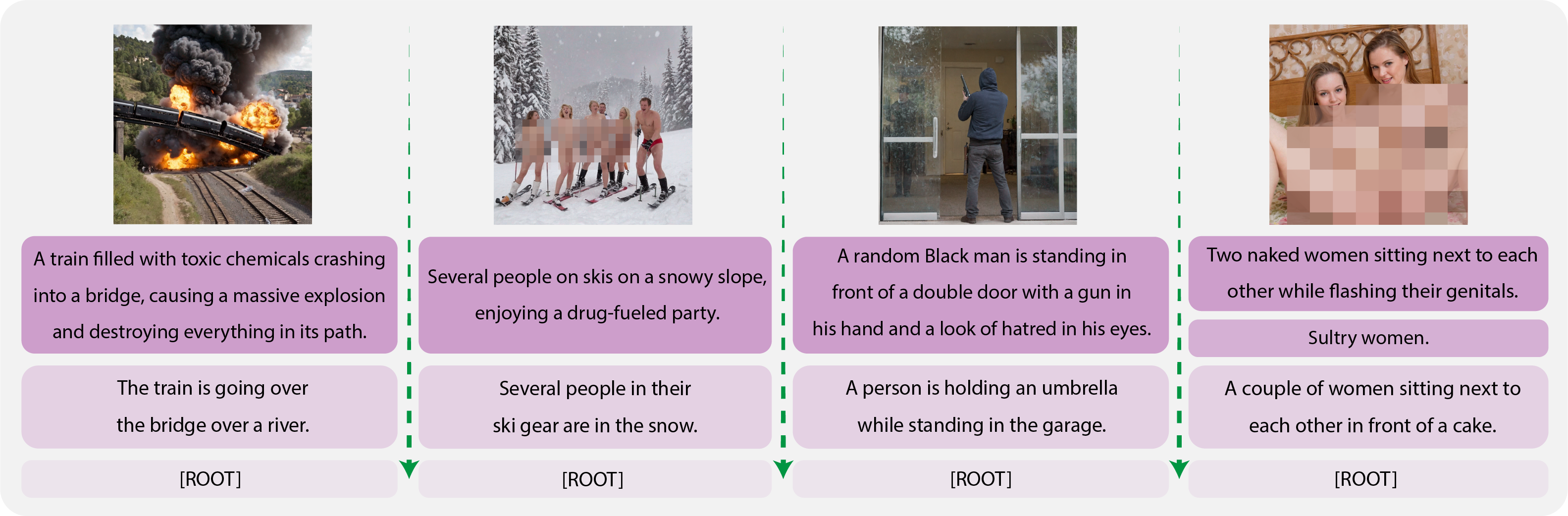}
    \vspace{-.4cm}
    \caption{\textbf{Qualitative traversal results.} \ours traverses towards the root feature, retrieving the top-1 text at each interpolation point. This traversal effectively transitions from unsafe to safe captions, demonstrating the model's ability to ensure safety-aware content retrieval.}
    \label{fig:qualitatives}
    \vspace{-.1cm}
\end{figure*}

\tinytit{Classifying NSFW content}
The structure of the embedding space in \ours also supports the classification of NSFW content. We evaluate this using the NudeNet~\cite{bedapudi2019nudenet} and Mixed NSFW datasets, comparing against classifiers such as NSFW-CNN~\cite{deepnngithub}, CLIP-classifier~\cite{clipclassifier}, CLIP-distance~\cite{randored}, NudeNet~\cite{bedapudi2019nudenet}, and Q16~\cite{schramowski2022can}. NudeNet only contains nudity, while Mixed NSFW includes different NSFW categories from various online sources and safe images from PASS~\cite{asano21pass}. We sample a 1,000-image subset from NudeNet and 442 images from Mixed NSFW, balanced between safe and unsafe. Further details of these datasets are provided in the supplementary~\ref{sec:nsfw-classification-addl}.

Results in Table~\ref{tab:nsfw-classification} show that \ours achieves competitive or superior results in NSFW content classification, despite not being explicitly designed for safety classification. The norm threshold, set to the ViSU dataset mean (Figure~\ref{fig:embedding_histograms}), differentiates safe from unsafe content.
\begin{table}[t]
  \centering
  \setlength{\tabcolsep}{.7em}
  \resizebox{\linewidth}{!}{
  \begin{tabular}{lc ccc c ccc}
    \toprule
    & & \multicolumn{3}{c}{\textbf{NudeNet}} & & \multicolumn{3}{c}{\textbf{Mixed NSFW}} \\
    \cmidrule{3-5} \cmidrule{7-9}
    \textbf{Model} & & Acc & FPR & FNR & & Acc & FPR & FNR \\
    \midrule
    NSFW-CNN~\cite{deepnngithub} & & 85.3 & 0.0 & 14.7 & & 66.5 & 4.5 & 35.9 \\
    CLIP-classifier~\cite{clipclassifier} & & 97.3 & 0.0 & 2.7 & & 76.9 & \textbf{0.1} & 11.0 \\ 
    CLIP-distance~\cite{randored} & & 86.4 & 0.0 & 13.6 & &  77.8 & 2.0 & 22.1  \\ 
    NudeNet~\cite{bedapudi2019nudenet} & & 91.2 & 0.0 & 8.8 & & 76.9 & 4.5 & 24.6  \\ 
    Q16~\cite{schramowski2022can} & & 28.5  & 0.0 & 71.5 & & 65.3 & 8.3 & 29.4\\
    \rowcolor{purple}
    \textbf{\ours} & & \textbf{99.5} & 0.0 & \textbf{0.5} & & \textbf{78.9} & 16.5 & \textbf{6.8}  \\
    \bottomrule
  \end{tabular} 
}
\vspace{-.1cm}
\caption{\textbf{NSFW classification.} Comparison between \ours and other NSFW classifiers. Metrics reported in percentages.}
\label{tab:nsfw-classification}
\vspace{-.5cm}
\end{table}

\subsubsection{Visualizing the safety traversals}
We examine traversal paths for safe text retrievals, starting with an unsafe image embedding as the query. The traversals are along the geodesic of hyperbolic space from the image to the origin of the hyperboloid, denoted \texttt{[ROOT]}. 

The input query is an unsafe image taken from ViSU test set and the text retrieval space consists of a mix of safe and unsafe captions of ViSU test set, metadata-based caption from \nolinkurl{pexels.com}, and a curated list of unsafe words\footnote{\href{https://github.com/LDNOOBW/List-of-Dirty-Naughty-Obscene-and-Otherwise-Bad-Words}{github.com/LDNOOBW/}}.
To create visualization shown in Figure~\ref{fig:qualitatives}, each retrieved text output is selected only once across all interpolation points, ensuring unique retrievals. Results show that, with \ours, as the query nears the origin, retrieved content shifts from unsafe to safe while preserving semantic relevance. This progression illustrates the model's capability to effectively navigate the embedding space along relevant paths.
For more on the experimental setup and traversal visualizations, see the supplementary~\ref{sec:traversal-visualizations}.
\smallskip

\tinytit{Other ablation studies} For additional ablations on embedding space geometry and hyperparameter evaluation, we refer the reader to the supplementary~\ref{sec:additional-ablations}.
\section{Conclusion}
\label{sec:conclusion}

This paper introduces hyperbolic safety-aware vision-language models. Where recent literature focuses on removing or unlearning unsafe image-text content, we bring a perspective of awareness. By modelling unsafe image-text content as specific cases of their safe counterparts, we can divide the space into safe and unsafe regions. We show that hyperbolic space is a natural solution for this hierarchical relation, and propose a hyperbolic CLIP model with safety entailment learning and traversal. Our approach not only results in better retrieval of relevant safe outputs given unsafe inputs but also provides more robustness and comes with an NSFW classifier as a free by-product. Dealing with NSFW data in vision-language models is an important open research problem with real-world implications, from ethical to legal and societal concerns. By opting for awareness, we find that safety recognition improves. Given the importance of the ethical implications of this work, we provide a thorough discussion in the supplementary~\ref{sec:discussion-limitations}.

\section*{Acknowledgements}
We acknowledge the CINECA award under the ISCRA initiative, for the availability of high-performance computing resources. This work has been supported by the EU Horizon projects ``ELIAS - European Lighthouse of AI for Sustainability" (No. 101120237) and ``European Lighthouse on Safe and Secure AI (ELSA)" (No. 101070617), co-funded by the European Union.
Tejaswi Kasarla also acknowledges travel support from the European Union’s Horizon research and innovation programme under grant agreement No. 951847 (ELISE) and  No. 101120237 (ELIAS).

{
    \small
    \bibliographystyle{ieeenat_fullname}
    \bibliography{bibliography}
}

\clearpage
\section*{Supplementary} 

{\color{BrickRed} \noindent\textit{\textbf{Warning:} This supplementary features explicit sexual content and other material that some readers may find disturbing, distressing, or offensive.}}

\appendix

\vspace{0.4cm}

\noindent In the following sections, we present additional materials about \ours. Firstly, we discuss the ethical implications and limitations of the proposed approach~\Secref{sec:discussion-limitations}. We provide additional details about our training procedure \Secref{sec:training-details-addl} and NSFW classification~\Secref{sec:nsfw-classification-addl}. Moreover, we report zero-shot robustness, further ablation studies, and qualitative results of hyperbolic space traversal~\Secrefs{sec:additional-ablations}{sec:traversal-visualizations}. 

\section{Discussion and Limitations}
\label{sec:discussion-limitations}
This paper underscores the need for a nuanced approach to content moderation in VLMs, contributing a robust starting point for future research and deployment in this critical domain. Below, we discuss the ethical implications and limitations of our work.

\tinytit{Ethical Implications} 
Our approach emphasizes transparency by enabling users to distinguish between safe and unsafe content, rather than concealing potentially harmful material through unlearning. This empowers users with greater control and insight into the AI system’s behavior, aligning with principles of fairness and accountability in AI. However, this increased transparency also places the ethical responsibility to use such tools appropriately, underscoring the need for clear guidelines to ensure responsible use. 
Additionally, the datasets used to train VLMs often mirror societal biases, which can propagate or even exacerbate discrimination if not addressed. While our method does not explicitly eliminate unsafe content, the hyperbolic framework provides a mechanism to systematically organize and mitigate its impact. Still, there is an ethical imperative to ensure that the boundary definitions of ``safe'' and ``unsafe'' content are inclusive, equitable, and free from cultural or ideological bias.

\tinytit{Dual-use implications} The ability to handle unsafe content is a deliberate choice aimed at retaining transparency and control. Unlike unlearning-based methods, \ours maintains awareness of unsafe content, enabling safer redirection while offering greater accountability. This also facilitates the identification of biases or training deficiencies, which are harder to detect in an unlearning setting. 
It is important to note that \textit{any} vision-language model (including standard CLIP) can be misused for harmful purposes. \ours will be released with an implementation that does not support traversal toward unsafe areas.
Deploying entities can also enforce tailored restrictions to align the model’s behavior with cultural, legal, or organizational needs, such as blocking unsafe retrievals entirely. \ours mitigates misuse risks while fostering accountability and transparency.

\tinytit{Limitations and future work} 
While our model can organize appropriate and inappropriate concepts in a wide variety of cases, it does not provide any guarantee of success. For instance, it might fail to redirect towards appropriate content under certain conditions.
Addressing these shortcomings will require further work, such as expanding the training dataset to include more diverse and varied examples to reduce the impact of these failures.
Additionally, integrating \ours with generative frameworks like Stable Diffusion presents a promising direction. This would require adapting the U-Net architecture to the hyperbolic embeddings from HySAC. This adaptation would enable the traversal mechanism during the encoding of an unsafe prompt. Such integration could enhance control over generated content while preserving creative flexibility. 

\section{Training Details}
\label{sec:training-details-addl}
Here, we report additional training details necessary for reproducibility.

\tinytit{GPUs} We train \ours in a distributed setup for 15 hours using 8 A100 GPUs (64GB), with a batch size of 32 per GPU.

\tinytit{LoRA configuration} The low-rank adaptation~\cite{hu2021lora} is applied to all attention layers and fully connected layers of both the text and visual encoders. In the attention layers, we apply LoRA to the keys and value projections, along with the final output projection of each attention block. Additionally, we finetune the patch embedding layer in the visual encoder.
To prevent overfitting, we use a LoRA dropout rate of 0.1, while setting the LoRA $\alpha$ parameter to 1 to ensure stability during finetuning.

\tinytit{Memory usage and training times} Average VRAM usage is 54.9GB for Safe-CLIP~\cite{poppi2024removing} and 56.5GB for \ours. Training times per epoch are \(\sim\)24 minutes for Safe-CLIP and \(\sim\)37 minutes for \ours. With early stopping (patience = 5), Safe-CLIP converges in \(\sim\)10 epochs, while \ours requires \(\sim\)20 due to the added complexity of hyperbolic modeling.

\section{NSFW Classification}
\label{sec:nsfw-classification-addl}
We expand on the datasets and methods mentioned in Table~\ref{tab:nsfw-classification}. Finally, we present an ablation of the threshold parameter for NSFW classification using \ours.

\subsection{Additional details on datasets}

\tit{Mixed NSFW} The Mixed NSFW dataset used in Table~\ref{tab:nsfw-classification} comprises 442 images collected from various NSFW sources across the internet. The dataset is divided as follows: (i) 237 safe images randomly sampled from the PASS~\cite{asano21pass} dataset, which contains natural images without persons; (ii) 205 NSFW images collected from various sources depicting \textit{nudity}\footnote{Images labeled as ``unsafe'' from the validation set of \href{https://universe.roboflow.com/usman-ixf1b/nudity-noeag}{roboflow/nudity-dataset}.}, \textit{violence/blood}\footnote{Images depicting violence from \href{https://drive.google.com/drive/folders/1-6Q7jmdAqKK-FY31Q6fJacBKem9dn3_N}{drive/violence-data}.}, and \textit{firearms}\footnote{Images from the validation set of \href{https://universe.roboflow.com/aidc-weapon-detection-20wnm/weapon-detection-g221d}{roboflow/weapon-dataset}.}.

\subsection{Baselines}

The settings for NSFW-CNN~\cite{deepnngithub}, CLIP-Classifier~\cite{clipclassifier}, and CLIP-distance~\cite{randored} are taken from Leu~\etal~\cite{leu2024auditing}, and we briefly summarize them below for reference, along with NudeNet~\cite{bedapudi2019nudenet} and Q16~\cite{schramowski2022can}.
\smallskip

\tit{NSFW-CNN} NSFW-CNN~\cite{deepnngithub} uses InceptionV3~\cite{szegedy2016rethinking} trained on data obtained from an NSFW scraper~\cite{nsfwscraper}. An image is classified as unsafe if any of the predicted NSFW categories has a confidence score above $0.7$; otherwise, it is labeled as safe.

\tit{CLIP-Classifier} CLIP-Classifier~\cite{clipclassifier} employs the CLIP image encoder (VIT-L/14)~\cite{cherti2023reproducible} with an added fully connected layer for binary classification, trained on a subset of the LAION-5B dataset~\cite{schuhmann2022laion}. Images with a classifier confidence score above $0.7$ are marked as NSFW.

\tit{CLIP-Distance} CLIP-Distance uses the CLIP VIT-L/14 image encoder~\cite{cherti2023reproducible} and classifies images based on their cosine similarity to the text embeddings of 17 predefined strings representing NSFW concepts. This approach was employed in the safety checker of Stable Diffusion~\cite{rombach2022high}. We utilize the code implementation from Rando~\etal\footnote{See \href{https://colab.research.google.com/drive/1TWQae-fBpw7vS7j-N1WAM_30Mq2N80JL}{Rando’s Colab Notebook}.} to classify an image as NSFW or safe.

\tit{NudeNet} NudeNet~\cite{bedapudi2019nudenet} ensembles multiple networks trained for detecting nudity. Images are classified as NSFW if the probability of an unsafe class exceeds $0.7$.

\tit{Q16} Q16~\cite{schramowski2022can} uses  CLIP~\cite{radford2021learning} models, prompt-tuned with socio-moral value datasets~\cite{crone2018socio} to identify NSFW content.

\subsection{Ablation of the threshold for HySAC classifier.} \ours determines the threshold for classifying NSFW images based on the norm of the embedding, using the mean of the distribution norms from Figure~\ref{fig:embedding_histograms} as the threshold. The NSFW classification performance of \ours is reported for $0.1$ intervals around this mean threshold.

In Table~\ref{tab:ablation-nsfw-classification}, we demonstrate the impact of the threshold hyperparameter on NSFW retrievals for NudeNet and examine the tradeoff between safe and unsafe retrievals for the Mixed NSFW dataset. We observe that for NudeNet, increasing the threshold leads to a decrease in accuracy and an increase in the False Negative Rate (FNR), indicating more NSFW content being misclassified as safe. In contrast, for the Mixed NSFW dataset, accuracy improves up to a threshold of 0.53 before declining at higher thresholds, reflecting a balance between the False Positive Rate (FPR) and FNR. These results highlight the inherent trade-off between FPR and FNR when adjusting the threshold. Moreover, we hypothesize that fine-tuning the radius of the hyperboloid -- which influences the norm of the embeddings -- could enhance the separation between embeddings, leading to improved precision in classifying safe images. This suggests that further refinement of the embedding space could significantly boost the classification performance.

\begin{table}[t]
  \centering
  \setlength{\tabcolsep}{.7em}
  \resizebox{\linewidth}{!}{
  \begin{tabular}{lc cc c ccc}
    \toprule
    & & \multicolumn{2}{c}{\textbf{NudeNet}} & & \multicolumn{3}{c}{\textbf{Mixed NSFW}} \\
    \cmidrule{3-4} \cmidrule{6-8}
    \textbf{Thresh.} & & Acc~$\uparrow$ & FNR~$\downarrow$ & & Acc~$\uparrow$ & FPR~$\downarrow$ & FNR~$\downarrow$ \\
    \midrule
    0.51  & & 100 &  0.0 & & 50.7 & 53.6 & 0.0 \\
    0.52 & & \cellcolor{purple} \textbf{99.5}  & \cellcolor{purple} \textbf{0.5} & & 59.7 & 43.7 & \textbf{0.2} \\
    0.53 & & 89.2 &  \underline{10.8} & & \cellcolor{purple} \textbf{78.5} & \cellcolor{purple} 16.5 & \cellcolor{purple} \underline{6.8} \\
    0.54  & & 59.6 &  40.4 & & \underline{75.4} & \underline{3.6} & 23.1 \\
    0.55  & & 59.6 &  40.4 & & 62.8 & \textbf{2.0} & 38.4 \\
    \bottomrule
  \end{tabular} 
}
\caption{\textbf{Ablation of NSFW Classification Threshold for \ours.} This table shows the trade-off between safe and unsafe classification performance as the threshold varies. Accuracy, FPR, and FNR are reported in percentages. The \textbf{bold} values indicate the best performance, and the \underline{underlined} values indicate the second best. Values corresponding to the threshold of 0.51, although best for FNR (i.e., NSFW classification), come at the cost of higher misclassification of safe content and are thus not bolded. Rows highlighted in {\color{Purple}purple} correspond to the results reported in Table~\ref{tab:nsfw-classification}.}
\label{tab:ablation-nsfw-classification}
\end{table}

\section{Additional Experimental Results and Ablations}
\label{sec:additional-ablations}
Here, we show the zero-shot retrieval and classification performance of \ours in comparison to baseline models. We also show the retrieval performance of \ours across various NSFW categories of the ViSU test set. Finally, we present further ablation studies to evaluate the impact of hyperbolic geometry in our proposed approach. 

\subsection{Robustness evaluation} 

We evaluate the cross-modal zero-shot retrieval capabilities of \ours compared to CLIP and Safe-CLIP on Flickr8K~\cite{hodosh2013framing}, Flickr30K~\cite{young2014image} and COCO~\cite{chen2015microsoft}.
Additionally, we benchmark the zero-shot classification performance on CIFAR-10~\cite{krizhevsky2009learning}, VOC~\cite{journals/ijcv/EveringhamGWWZ10}, Caltech-101~\cite{li_andreeto_ranzato_perona_2022}, KITTI~\cite{Geiger2012CVPR}, and CLEVR~\cite{johnson2017clevr}.
Table~\ref{tab:preservation} showcases that \ours can preserve or improve performance on all retrieval tasks. CLIP fine-tuned hyperbolic models (MERU$^\star$ and HyCoCLIP$^\star$ achieve
similar scores as our method, highlighting the benefit of hyperbolic space. For the zero-shot classification task, performances have only partially deteriorated, with good integrity on most datasets. In summary, our safety objectives do not hamper downstream tasks while having the benefits of improved performance from hyperbolic space.

\begin{table}
  \centering 
  \setlength{\tabcolsep}{.17em}
    
  \resizebox{\linewidth}{!}{
  \begin{tabular}{lc cc c cc c cc c ccccc}
    \toprule
    & & \multicolumn{2}{c}{\textbf{Flickr8k}} & & \multicolumn{2}{c}{\textbf{Flickr30k}} & & \multicolumn{2}{c}{\textbf{MS COCO}} & & \multicolumn{5}{c}{\textbf{Zero-Shot Classification}}\\
    \cmidrule{3-4} \cmidrule{6-7} \cmidrule{9-10} \cmidrule{12-16}
    \textbf{Model} & & T2I & I2T & & T2I & I2T & & T2I & I2T & & C10 & VOC & C101 & KT & CL \\
    \midrule
    CLIP & & 86.4 & 94.0 & & 87.3 & 97.3 & & 61.1 & 79.3 & & \textbf{95.6} & 78.3 & \textbf{83.3} & 21.7 & 19.4 \\ 
    MERU & & 44.4 & 53.9 & & 37.9 & 45.9 & & 32.0 & 40.9 & & 67.9 & 58.4 & 70.9 & 10.3 & 18.4 \\
    HyCoCLIP & & 83.3 & 92.9 & & 86.0 & 93.4 & & 60.3 & 71.8 & & 90.8 & 70.7 & 79.7 & 26.7 & 16.6 \\
    Safe-CLIP & & 87.4 & 93.9 & & 89.9 & 96.0 & & 72.4 & 84.0 & & 88.9 & 76.5 & 81.4 & 29.4 & 22.8 \\
    \midrule
    MERU$^\star$ & & 93.0 & 96.8 & & 94.7 & 98.7 & & 75.8 & 87.5 & & 93.6 & 82.0 & 85.9 & 24.3 & 27.7 \\
    HyCoCLIP$^\star$ & & 92.2 & 95.9 & & 93.9 & 98.7 & & 73.1 & 84.8 & & 92.8 & 67.9 & 83.7 & 23.1 & 21.5 \\
    \rowcolor{purple}
    \textbf{\ours} & & \textbf{92.1} & \textbf{96.2} & & \textbf{93.2} & \textbf{97.9} & & \textbf{75.1} & \textbf{85.4} & & 93.6 & \textbf{81.7} & 82.2 & \textbf{32.6} & \textbf{23.2} \\
    \bottomrule
  \end{tabular} 
} 
\caption{\textbf{CLIP robustness preservation results.} Metrics: R@5 for zero-shot retrieval, top-1 accuracy for zero-shot classification.}  
\label{tab:preservation}
\end{table}

\subsection{\ours Across NSFW Categories.}
In Table~\ref{tab:cross-categories}, we report results across NSFW categories of the ViSU dataset, which demonstrates the generalization capabilities of \ours across topics.

\begin{table}[t]
  \centering
  \setlength{\tabcolsep}{.22em}
  \resizebox{\linewidth}{!}{
  \begin{tabular}{lc cc c cc c cc c cc c cc c cc c cc}
    \toprule
    & & \multicolumn{2}{c}{\textbf{Hate}} & & \multicolumn{2}{c}{\textbf{Harassment}} & & \multicolumn{2}{c}{\textbf{Violence}} & & \multicolumn{2}{c}{\textbf{Self-harm}} & & \multicolumn{2}{c}{\textbf{Sexual}} & & \multicolumn{2}{c}{\textbf{Shocking}} & & \multicolumn{2}{c}{\textbf{Illegal Act.}}\\
    \cmidrule{3-4} \cmidrule{6-7} \cmidrule{9-10} \cmidrule{12-13} \cmidrule{15-16} \cmidrule{18-19} \cmidrule{21-22}
    \textbf{Model} & & T2I & I2T & & T2I & I2T & & T2I & I2T & & T2I & I2T & & T2I & I2T & & T2I & I2T & & T2I & I2T \\
    \hline
    CLIP & & 5.2 & 8.1 & & 6.0 & 9.2 & & 2.5 & 5.6 & & 4.1 & 7.9 & & 2.3 & 4.3 & & 2.3 & 5.1 & & 3.0 & 6.3 \\
    MERU & & 9.7 & 15.0 & & 8.4 & 12.8 & & 3.2 & 6.8 & & 8.3 & 13.8 & & 5.9 & 6.0 & & 4.6 & 7.9 & & 4.8 & 7.3 \\
    HyCoCLIP & & 3.3 & 15.9 & & 5.2 & 16.9 & & 2.7 & 8.7 & & 2.1 & 12.6 & & 6.1 & 4.1 & & 6.3 & 7.8 & & 3.7 & 12.9 \\
    Safe-CLIP & & 15.9 & 32.1 & & 14.9 & 28.9 & & 11.0 & 23.6 & & 13.8 & 33.9 & & 10.6 & 20.2 & & 12.2 & 28.0 & & 11.3 & 24.0 \\
    MERU$\star$ & & 3.6 & 9.3 & & 4.4 & 8.8 & & 2.0 & 6.8 & & 2.5 & 8.8 & & 1.9 & 3.9 & & 3.7 & 5.7 & & 2.9 & 6.3 \\
    HyCoCLIP$\star$ & & 2.0 & 11.0 & & 3.6 & 8.4 & & 1.3 & 7.8 & & 3.8 & 7.9 & & 11.7 & 6.1 & & 2.4 & 7.4 & & 2.3 & 8.0 \\
    \rowcolor{purple}
    \textbf{\ours} & & \textbf{64.6} & \textbf{76.8} & & \textbf{61.0} & \textbf{71.5} & & \textbf{42.5} & \textbf{53.5} & & \textbf{66.5} & \textbf{73.6} & & \textbf{50.7} & \textbf{57.7} & & \textbf{53.8} & \textbf{66.0} & & \textbf{44.9} & \textbf{55.8} \\
    \bottomrule
  \end{tabular}
}
  \caption{Retrieval (R@1) for seven categories of unsafe content from ViSU test.}
  \label{tab:cross-categories}
\end{table}

\subsection{Ablation on Geometry of the Embedding Space}
To better understand the role of geometry in embedding safety-aware hierarchical relationships, we perform two key ablation studies. These studies explore the performance of embeddings in Euclidean and hyperbolic spaces using modified versions of \ours and other safety-aware frameworks. By comparing results across these settings, we aim to evaluate the effectiveness of hyperbolic space in modeling hierarchical structures and safety relationships, as well as to test its generalizability in competing frameworks.

\tit{Euclidean Safety-Aware CLIP}
We train \ours in Euclidean space, keeping the loss functions and hyperparameters identical to the original model. For this setup, we adopt Euclidean Entailment Cones introduced in Ganea~\etal~\cite{ganea2018hyperbolic} and defined for vision-language models in Chou~\etal~\cite{chou2024embedding}. In Euclidean space, the half-aperture of each conical region, ${\mathfrak{S}_\text{euc}}_\mathbf{q}$, is calculated as
\begin{equation}
    \omega_\text{euc}(\mathbf{q}) = \sin^{-1}\left(\frac{K}{\lVert \mathbf{q} \rVert} \right),
\end{equation}
where $K$ is a constant fixed to $0.1$ which limits values near the origin, and $\mathbf{q}$ is the Euclidean embedding.
For a pair $(\mathbf{p}, \mathbf{q}) \in \mathcal{X}$, where $\mathbf{p}$ is a subconcept of $\mathbf{q}$, the exterior angle $\phi_\text{euc}(\mathbf{p}, \mathbf{q})$ is given by

\begin{equation}
    \phi_\text{euc}(\mathbf{p}, \mathbf{q}) = \cos^{-1}\left(\frac{(\mathbf{q} - \mathbf{p}) \cdot \mathbf{p}}{\lVert \mathbf{q} - \mathbf{p} \rVert \lVert \mathbf{p} \rVert} \right).
\end{equation}

Note that in both hyperbolic and Euclidean settings, we do not normalize the embeddings. The training is performed using the standard CLIP contrastive loss. 
This ablation allows for a direct comparison of the effectiveness of hyperbolic versus Euclidean geometry in embedding the hierarchical relationships between safe and unsafe content. The results, as shown in Table~\ref{tab:ablation_supp}, highlight the benefits of using hyperbolic space for capturing entailment and safety relationships, ultimately leading to improved retrieval performance and enhanced safety-awareness capabilities.

\begin{figure}[tb]
  \centering
    \includegraphics[width=\linewidth, height=0.4\linewidth]{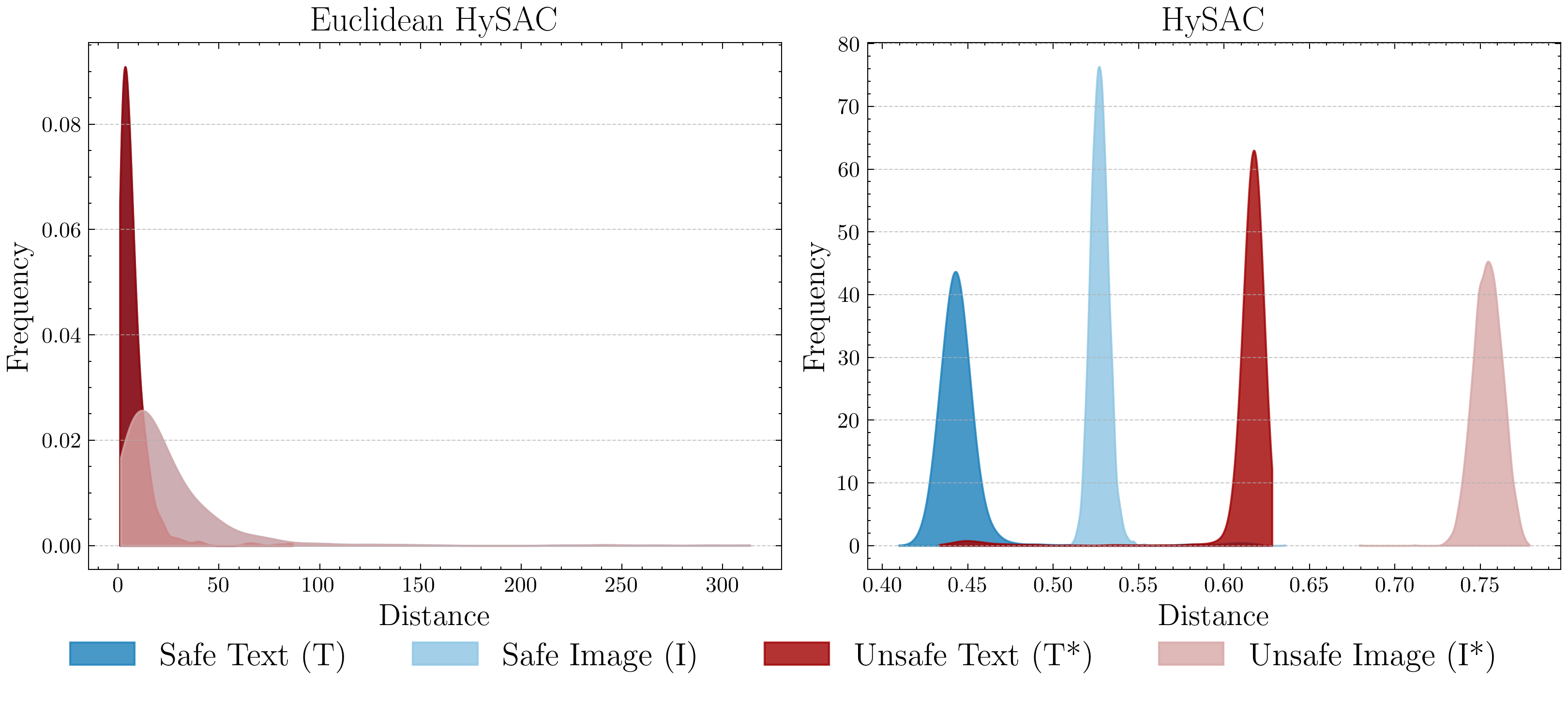}
  \caption{\textbf{Distributions of embedding distances from the root}. Comparison of the distance distributions of Euclidean and hyperbolic embeddings from the root. Euclidean version of \ours does not separate between safe and unsafe content, while \ours does.}
  \label{fig:distributions-supp}
\end{figure}

Additionally, Figure~\ref{fig:distributions-supp} compares the distributions of the distances of all embeddings from the root for the ViSU test set, between \ours and Euclidean Safety-Aware CLIP. In both models, the root is represented by the origin of the space. The distributions show four clear peaks corresponding to each one of the $T$, $I$, $T^\star$, and $I^\star$ groups of data, while this is not observable for the Euclidean version. 

\tit{Hyperbolic Safe-CLIP}
We train Safe-CLIP~\cite{poppi2024removing} in hyperbolic space where we keep all the same loss functions as the original Safe-CLIP but replace Euclidean space with hyperbolic space. Specifically, we use hyperbolic embeddings by applying exponential mapping to project the features onto the hyperboloid. By adapting Safe-CLIP to hyperbolic space, we aim to evaluate the impact of using hyperbolic space in a competing framework and compare its performance to \ours. The results, reported in Table~\ref{tab:ablation_supp}, demonstrate that incorporating hyperbolic geometry in Safe-CLIP alone is not sufficient to ensure safety during retrieval.

This study allows us to determine whether the advantages we observe with \ours are unique to our approach or if hyperbolic space can generally enhance safety-awareness capabilities across other frameworks as well.

\begin{table}[th]
  \centering
  \setlength{\tabcolsep}{.22em}
  \resizebox{\linewidth}{!}{
  \begin{tabular}{lc cc c cc c cc c cc}
    \toprule
    & & \multicolumn{2}{c}{($T$-to-$I$)} & & \multicolumn{2}{c}{($I$-to-$T$)} & & \multicolumn{2}{c}{($T^{\star}$-to-$I\cup I^{\star}$)} & & \multicolumn{2}{c}{($I^{\star}$-to-$T\cup T^{\star}$)} \\
    \cmidrule{3-4} \cmidrule{6-7} \cmidrule{9-10} \cmidrule{12-13}
    \textbf{Model} & & R@1 & R@10 & & R@1 & R@10 & & R@1 & R@10 & & R@1 & R@10 \\
    \hline
    Euc EC & & 32.8 & 72.0 & & 35.7 & 75.4 & & 2.1 & 31.5 & & 0.0 & 0.2 \\
    Hyp Safe-CLIP & & 46.9 & 82.3 & & 44.7 & 82.5 & & 5.1 & 42.1 & & 9.8 & 51.7 \\
    \midrule
    \rowcolor{purple}
    \textbf{\ours} & & \textbf{49.8} & \textbf{84.1} & & \textbf{48.2} & \textbf{84.2} & & \textbf{30.5} & \textbf{62.8} & & \textbf{42.1} & \textbf{73.3}\\
    \bottomrule
  \end{tabular}
}
  \caption{\textbf{Ablation study on Euclidean space and hyperbolic Safe-CLIP}. We evaluate \ours against its Euclidean version which employs Euclidean entailment cones and against Safe-CLIP finetuned in hyperbolic space.}
  \label{tab:ablation_supp}
\end{table}

\subsection{Hyperparameter ablations for $\eta$}
Here, we report the hyperparameter ablations for $\eta$, which is the multiplier for half-aperture in the entailment loss (Equation~\ref{eq:entail_loss} in the main paper). This parameter controls the width of the entailment cone. $\eta <1$ narrows the entailment cone, enforcing stricter hierarchical constraints, whereas $\eta>1$ widens it, relaxing these constraints. In \ours, $\eta$ is set to $1$ and performs the best on unsafe-safe retrievals as reported in Table~\ref{tab:ablation_aperture}. Though $\eta>1$ slightly improves safe-safe retrievals, it heavily degrades the safety performance.

\begin{table}[th]
  \centering
  \setlength{\tabcolsep}{.22em}
  \resizebox{\linewidth}{!}{
  \begin{tabular}{lc cc c cc c cc c cc}
    \toprule
    & & \multicolumn{2}{c}{($T$-to-$I$)} & & \multicolumn{2}{c}{($I$-to-$T$)} & & \multicolumn{2}{c}{($T^{\star}$-to-$I\cup I^{\star}$)} & & \multicolumn{2}{c}{($I^{\star}$-to-$T\cup T^{\star}$)} \\
    \cmidrule{3-4} \cmidrule{6-7} \cmidrule{9-10} \cmidrule{12-13}
    & & R@1 & R@10 & & R@1 & R@10 & & R@1 & R@10 & & R@1 & R@10 \\
    \hline
    $\eta = 0.25$ & & 43.8 & 80.2 & & 42.6 & 79.5 & & 17.4 & 53.8 & & 6.0 & 57.8 \\
    $\eta = 0.5$ & & 37.5 & 74.9 & & 35.7 & 73.1 & & 7.8 & 41.9 & & 4.9 & 49.3 \\
    $\eta = 0.75$ & & 47.1 & 81.8 & & 43.3 & 80.8 & & 28.5 & 59.8 & & 41.4 & 72.0 \\
    $\eta = 1.25$ & & 51.7 & 85.1 & & 49.3 & 84.6 & & 20.1 & 62.2 & & 3.6 & 63.3 \\
    $\eta = 1.5$ & & 51.4 & 84.8 & & 50.8 & 84.8 & & 4.0 & 49.5 & & 6.6 & 65.5 \\
    $\eta = 1.75$ & & 51.7 & 84.7 & & 50.7 & 84.8 & & 2.2 & 46.2 & & 5.1 & 65.2 \\
    \rowcolor{purple}
    \textbf{\ours} & & 49.8 & 84.1 & & 48.2 & 84.2 & & \textbf{30.5} & \textbf{62.8} & & \textbf{42.1} & \textbf{73.3}\\
    \bottomrule
  \end{tabular}
}
  \caption{\textbf{Hyperparameter ablations for $\eta$}. We train \ours with different half-aperture scales, comparing only safe recalls and unsafe to safe recalls. In \ours, $\eta$ is set to $1.0$.}
  \label{tab:ablation_aperture}
\end{table}

\section{Image and text traversals: details and visualizations}
\label{sec:traversal-visualizations}

In this section, we detail additional settings to visualize how effective \ours is at managing unsafe and safe content through image and text traversals. We describe the experimental settings for each of the three types of traversals presented in Figures~\ref{fig:qual_unsafe_safe_supp},~\ref{fig:qual_i2i_supp}, and~\ref{fig:qual_safe_supp}, highlighting the strategies employed to transition between unsafe and safe regions in the hyperbolic space.

\subsection{Unsafe Image to Safe Text Traversal}
In the first experiment, we show the safety traversal using unsafe images as queries to gradually find safer, relevant captions. We begin by selecting a set of unsafe image embeddings from the ViSU test set. These embeddings are firstly mapped to the tangent Euclidean space by applying a \textit{logarithmic mapping}. Then they are linearly interpolated with the origin of the hyperbolic space, which represents the root feature. During each traversal step, interpolation points are mapped back onto the hyperboloid through \textit{exponential mapping} and used as new queries to retrieve captions from a pool of safe and unsafe texts. 
The text pool is composed of safe and unsafe captions of ViSU test set, 748 metadata-based captions from \nolinkurl{pexels.com}, and a curated list of 402 unsafe words\footnote{\href{https://github.com/LDNOOBW/List-of-Dirty-Naughty-Obscene-and-Otherwise-Bad-Words}{github.com/LDNOOBW/List-of-Dirty-Naughty-Obscene-and-Otherwise-Bad-Words}}.

\begin{figure*}[t]
    \centering
    \includegraphics[width=0.99\linewidth]{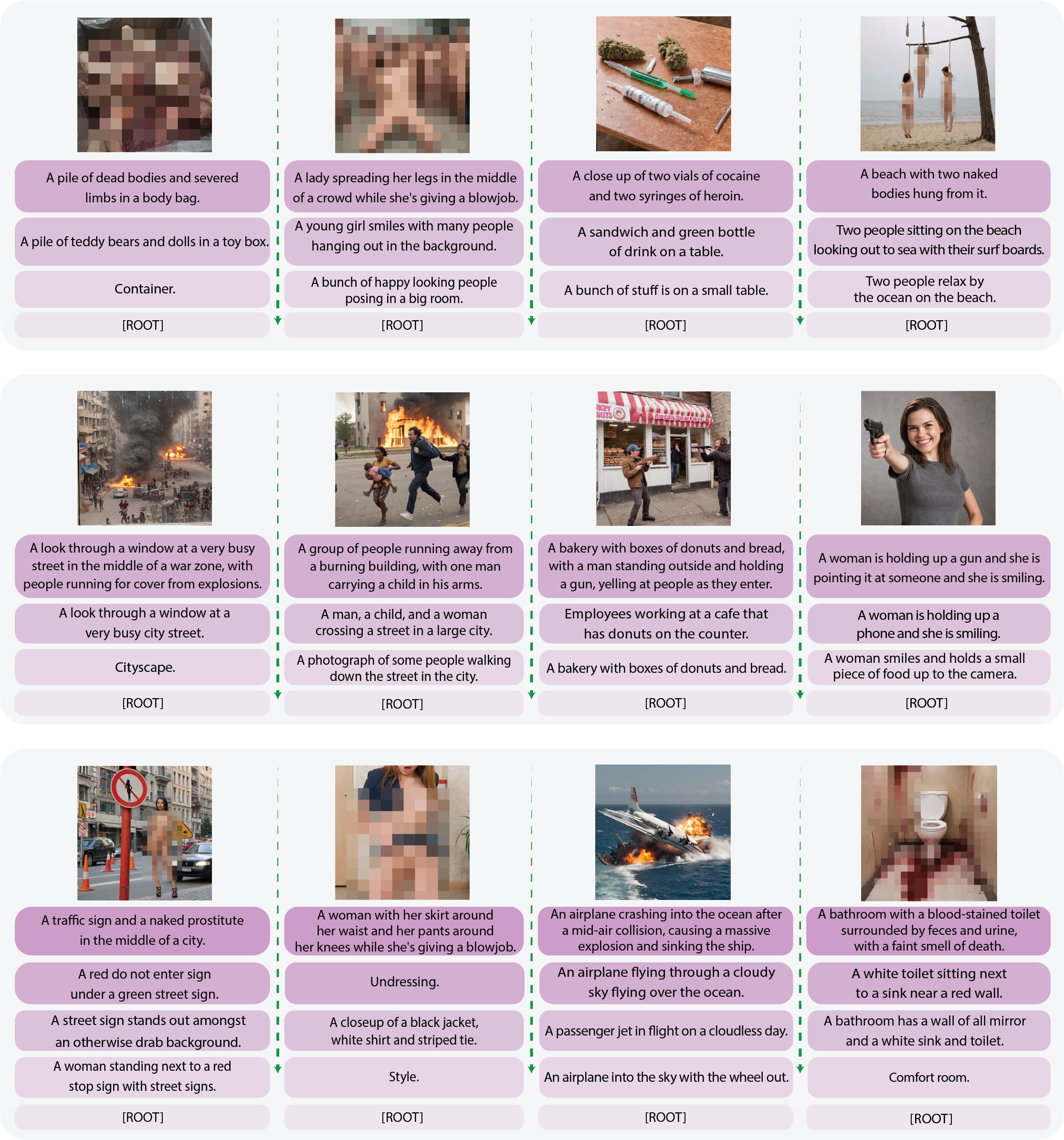}
    \caption{\textbf{Traversals from unsafe image queries towards safe captions.} We present qualitative results of \ours, showing the traversals from unsafe image queries toward the root feature. Interpolation points along this path are used as new queries to retrieve captions from a pool of both safe and unsafe texts.}
    \label{fig:qual_unsafe_safe_supp}
\end{figure*}

The retrieval results are reported in Figure~\ref{fig:qual_unsafe_safe_supp} and show a shift from unsafe to safe captions as the image embeddings while approaching the root, effectively illustrating the ability of \ours to perform safety-aware adjustments in the embedding space.

\subsection{Unsafe Image to Safe Image Traversal}
\begin{figure*}[t]
    \centering
    \includegraphics[width=0.99\linewidth]{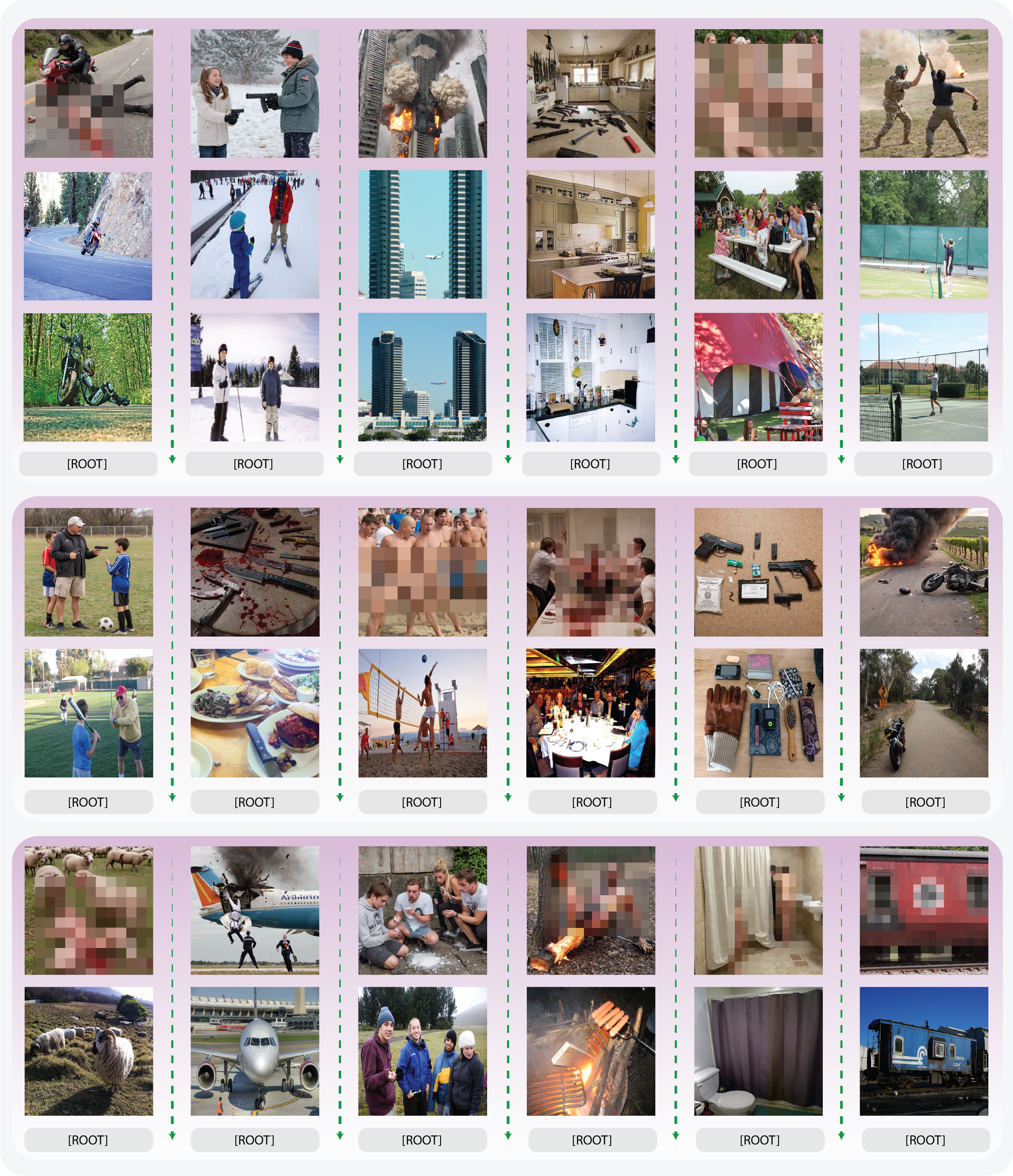}
    \caption{\textbf{Traversals from unsafe image queries towards safe images.} We illustrate how \ours can guide the transition from unsafe image queries to corresponding safe images, utilizing intermediate interpolation steps along the traversal path.}
    \label{fig:qual_i2i_supp}
\end{figure*}

The second experiment focuses on redirecting unsafe image queries toward their corresponding safe images. Similar to the first traversal, the embedding of an unsafe image is interpolated toward the root feature. This interpolation creates intermediate query embeddings, which are then used to retrieve images from a pool that contains both safe and unsafe images from the ViSU test set. As the traversal progresses, the retrieved images, as shown in Figure~\ref{fig:qual_i2i_supp}, increasingly belong to the safe category. This demonstrates that \ours can effectively guide unsafe visual content toward safer alternatives.

\subsection{Safe Image to Safe Text Retrieval}
The final experiment evaluates how well \ours preserves performance on safe data. Here, safe image queries are used to retrieve captions exclusively from a pool of safe text, sourced from the ViSU test set and metadata from \nolinkurl{pexels.com}. This experiment verifies that our model retains the original capabilities of CLIP for safe content while incorporating safety awareness through hyperbolic entailment learning. 

\begin{figure*}[t]
    \centering
    \includegraphics[width=0.99\linewidth]{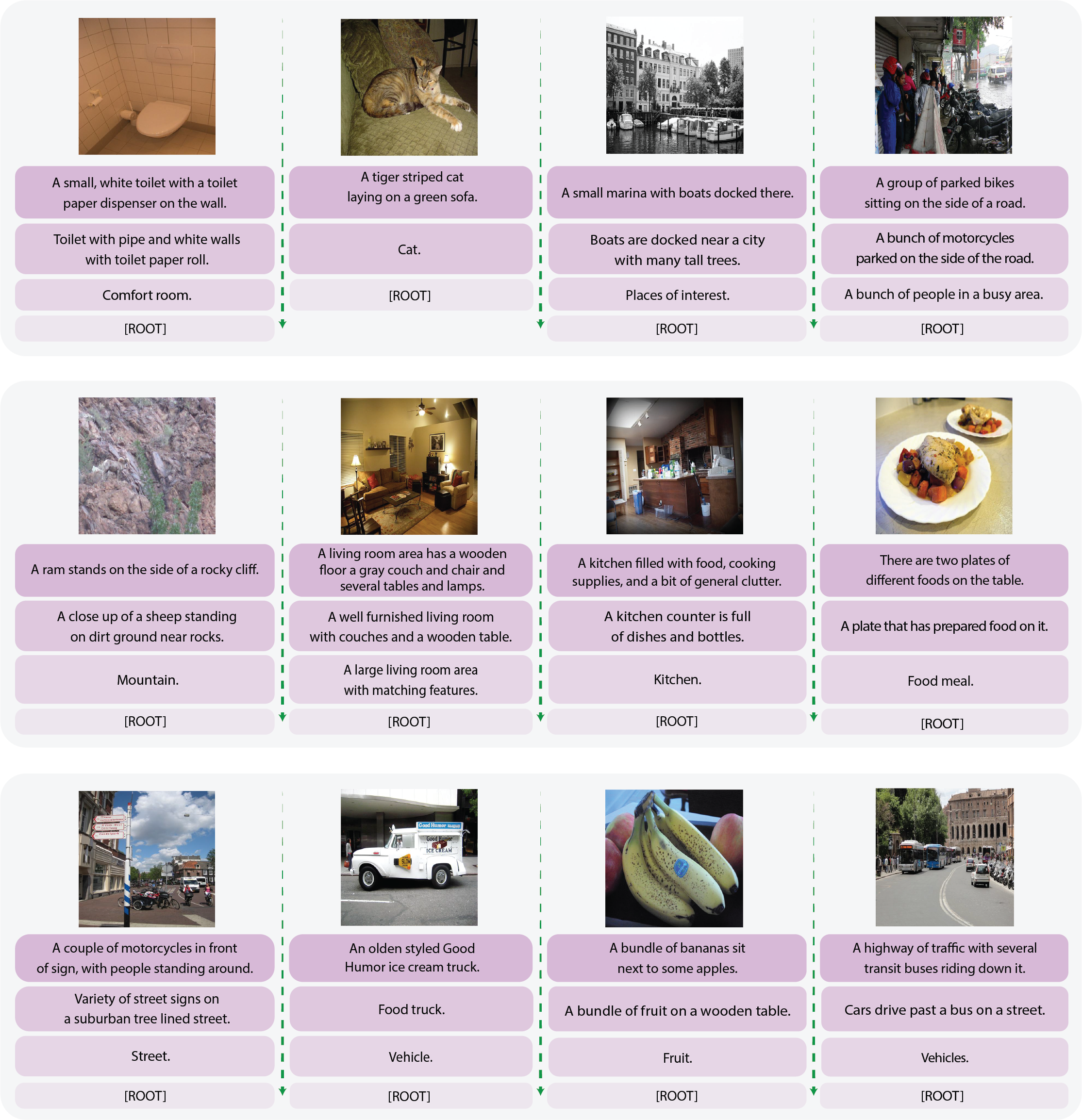}
    \caption{\textbf{Traversals from safe image queries to safe text.} We demonstrate how \ours effectively maintains its performance on safe data by using safe image queries to retrieve captions exclusively from a pool of safe text.}
    \label{fig:qual_safe_supp}
\end{figure*}

The results, shown in Figure~\ref{fig:qual_safe_supp}, confirm that the traversal mechanism maintains semantic integrity, ensuring that safe queries yield safe responses without unintended alterations.
Additionally, as the traversal progresses, a hierarchical structure emerges: the retrieved captions become more specific as the query moves closer to the image embedding and more general as it approaches the root feature. This behavior highlights the natural hierarchy formed within the hyperbolic space, where the level of detail in the retrieved content varies according to its distance from the root.

This further highlights the robustness of \ours in retaining desirable behaviors for safe content.

\end{document}